\pdfoutput=1

\documentclass[11pt]{article}

\usepackage[preprint]{acl}

\usepackage{times}
\usepackage{latexsym}

\usepackage[T1]{fontenc}

\usepackage[utf8]{inputenc}

\usepackage{microtype}

\usepackage{inconsolata}

\usepackage{microtype}
\usepackage{graphicx}
\usepackage{booktabs}
\usepackage{makecell}
\usepackage{multirow}
\usepackage{wrapfig}
\usepackage{xspace}
\usepackage{subcaption}
\usepackage{hyperref}
\usepackage{xcolor}
\usepackage{algorithm}
\usepackage{algpseudocode}

\newcommand{\scenedataname}{\textsc{DivScene}\xspace}
\newcommand{\modelname}{\textsc{NatVLM}\xspace}
\newcommand{\trajdataname}{\textsc{DivScene}$_{ep}$\xspace}
\newcommand{\moveahead}{\textsc{MoveAhead}\xspace}
\newcommand{\rotateright}{\textsc{RotateRight}\xspace}
\newcommand{\rotateleft}{\textsc{RotateLeft}\xspace}
\newcommand{\done}{\textsc{Done}\xspace}

\definecolor{inputcolor}{HTML}{e65100}
\definecolor{outputcolor}{HTML}{6a1b9a}

\usepackage{amsmath}
\usepackage{amssymb}
\usepackage{mathtools}
\usepackage[inline]{enumitem}
\usepackage{amsthm}
\usepackage[capitalize,noabbrev]{cleveref}

\title{\scenedataname: Towards Open-Vocabulary Object Navigation with Large Vision Language Models in Diverse Scenes}

\author{
  Zhaowei Wang$^1$, Hongming Zhang$^2$, Tianqing Fang$^{1,2}$, Ye Tian$^3$, Yue Yang$^4$, \\ \textbf{Kaixin Ma$^2$, Xiaoman Pan$^2$, Yangqiu Song$^1$, and Dong Yu$^2$} \\
  $^1$CSE Department, HKUST, $^2$Tencent AI Lab, Bellevue, USA\\ 
  $^3$Robotics X, Tencent,
  $^4$University of Pennsylvania\\
\texttt{zwanggy@cse.ust.hk, hongmzhang@global.tencent.com}
  }

\begin{document}
\maketitle
\begin{abstract}
Large Vision-Language Models (LVLMs) have achieved significant progress in tasks like visual question answering and document understanding. However, their potential to comprehend embodied environments and navigate within them remains underexplored.
In this work, we first study the challenge of open-vocabulary object navigation by introducing \scenedataname, a large-scale dataset with 4,614 houses across 81 scene types and 5,707 kinds of target objects. 
Our dataset provides a much greater diversity of target objects and scene types than existing datasets, enabling a comprehensive task evaluation.
We evaluated various methods with LVLMs and LLMs on our dataset and found that current models still fall short of open-vocab object navigation ability. Then, we fine-tuned LVLMs\footnote{Our code and data are available at \url{https://github.com/zhaowei-wang-nlp/DivScene}.} to predict the next action with CoT explanations. We observe that LVLM's navigation ability can be improved substantially with only BFS-generated shortest paths without any human supervision, surpassing GPT-4o by over 20\% in success rates.
\end{abstract}

\section{Introduction}
Large Vision-Language Models (LVLMs), such as Qwen2.5-VL~\cite{bai2025qwen2}, GPT-4o~\cite{2023GPT4VisionSC}, and others~\cite{zhu2025internvl3}, have demonstrated state-of-the-art performance on a variety of vision-and-language tasks. However, their ability to comprehend embodied environments and navigate within them has been less explored, primarily due to the limited diversity of scenes and objects in current navigation benchmarks. For example, Matterport-3D~\citep{chang2017matterport3d} only considers 21 types of target objects in 90 private homes, and ProcTHOR~\citep{deitke2022️} contains 16 object types in four kinds of rooms (i.e., bedroom, living room, kitchen, and bathroom).

\begin{figure}[t]
    \centering
  \includegraphics[width=0.95\columnwidth]{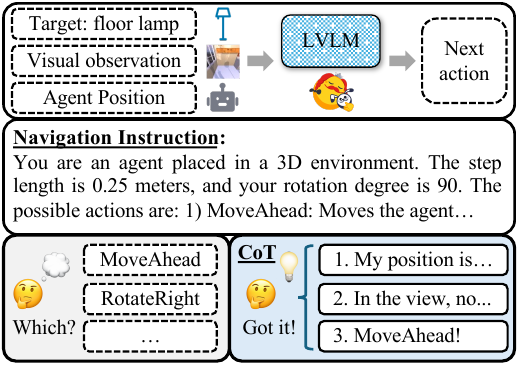}
  \caption{Illustration of our open-vocabulary object navigation. When fine-tuning LVLMs, we build CoT explanation traces to help them better grasp the rationale of object navigation.}
  \vspace{-0.45cm}
  \label{fig:intro_figure}
\end{figure}

In this work, we propose the new task of open-vocab object navigation and first study LVLMs' performance on the task, where an agent is required to navigate to a wide range of target objects without a pre-defined vocabulary. With this purpose, we introduce a new dataset, \scenedataname, which features the most comprehensive range of scene types and target objects to the best of our knowledge. Specifically, we collect 81 scene types based on the MIT Scenes Dataset~\citep{quattoni2009recognizing}.
Then, we use LLMs to automatically compose diverse house descriptions by adding attributes to those scene types, such as ``a \textit{bakery} with tile-patterned walls.'' We input these descriptions into a language-guided framework, Holodeck~\citep{yang2024holodeck}, to build houses automatically with the strong ability of GPT-4~\citep{2023GPT4VisionSC}. 
In total, we compile 4,614 houses across 81 distinct scene types on the AI2THOR platform~\citep{kolve2017ai2}. For benchmarking and training LVLMs, we further sampled shortest-path episodes in the houses from \scenedataname. Specifically, we discretize houses into grid maps with a fixed step size and randomly sample target objects in each house. Then, we search for the shortest paths from the agent's initial position to target objects with BFS. In total, we collect about 23K shortest-path episodes, forming the \trajdataname. In total, our episode data contains 5,707 types of target object, significantly surpassing existing object navigation datasets~\cite{deitke2020robothor, chang2017matterport3d}. This makes it a comprehensive testbed for evaluating the open-vocab navigation capabilities.

With our collected data, we benchmarked various methods based on LVLMs and LLMs, including blind LLMs, LLMs with image captioning, open-source LVLMs, and proprietary LVLMs. 
Our results show that most current models are still unable to explore environments and navigate toward open-vocabulary objects.
Specifically, most models failed to outperform a random baseline, while GPT-4o achieves a success rate of slightly above 30\%, far from adequate for real-world applications. 
To enhance their navigation ability, we further studied fine-tuning LVLMs on the data sampled in our \scenedataname. 
Specifically, we propose a new navigational LVLM, called \modelname (\textbf{Na}vigational Chain-of-\textbf{T}hought \textbf{VLM}), which is fine-tuned from Idefics 2~\citep{laurenccon2024matters} on our sampled episodes in \trajdataname. 
As shown in \cref{fig:intro_figure}, the model is tuned to process the current observation, such as the egocentric view and the agent's status, and generate the next action. 
We train Idefics 2 with imitation learning~\citep[IL;][]{brohan2022rt} using BFS-generated shortest paths in \trajdataname. 
To further improve the navigation ability, we also manually collect CoT explanation traces of each action prediction~\citep{mitra2023orca, ho2023large} to help Idefics 2 understand the underlying rationale behind the navigation task. 

In our experiments, we surprisingly discovered that simply imitating the shortest paths constructed by breadth-first search can be an effective approach to enhance the open-vocab navigation ability of LVLMs. Our \modelname model outperforms off-the-shelf LVLMs and LLMs by a large margin, achieving a success rate approximately 20\% higher than GPT-4o. Compared to existing IL-based methods, which rely on training models with large corpora of costly human demonstrations~\citep{brohan2022rt, wei2023imitation}, our approach offers a far more efficient and economical alternative. Further, we carry out thorough ablation studies to show the efficacy of CoT explanation traces in action prediction. 
Moreover, few-shot experiments demonstrate the robustness of our agent.
Last but not least, we validate the generalization ability of our agent on three unseen datasets: ProcTHOR~\citep{deitke2022️}, iTHOR~\citep{weihs2021visual}, and HM3D~\cite{ramakrishnan2021habitat}.



\section{Related Work}


\noindent\textbf{Object Navigation:}
The Reinforcement Learning has long been used for tackling object navigation tasks~\citep{zhu2017target, druon2020visual, ehsani2021manipulathor}. 
However, recent studies~\citep{ehsani2023imitating} show that RL requires extensive reward shaping, limiting its practicality.
Alternatively, imitation learning~\citep{pomerleau1988alvinn, zhang2018deep} has also been applied to these tasks, inspiring many subsequent works~\citep{brohan2022rt, brohan2023rt, ehsani2023imitating}. However, existing works only focus on closed-vocab object navigation tasks, mainly due to the limited diversity of current benchmarks. For example, Matterport-3D~\citep{chang2017matterport3d} only considers 21 kinds of target objects in private homes. ProcTHOR~\citep{deitke2023phone2proc} solely contains 16 target objects in four kinds of rooms.
Although HM3D-OVON~\citep{yokoyama2024hm3d} attempts to broaden the diversity, the number of object categories is still limited to 379, and there are only 181 scenes, making it difficult to fully reflect the open-vocab nature of real-world scenarios. 
In constrast, we introduce \scenedataname, a dataset comprising 5,707 categories of target objects among 22,696 object types, spanning 4,614 distinct scenes.
Therefore, our dataset reflects a more diverse nature of real-world settings with an open vocabulary.

\noindent\textbf{VLMs and LLMs for Embodiment:}
LLMs and VLMs~\citep{achiam2023gpt, touvron2023llama, lu2024deepseek} have emerged as the foundation for solving embodied tasks~\citep{pan2023langnav, majumdar2024openeqa}. A few methods have employed contrastive VLMs~\citep{radford2021learning, li2022grounded} as the visual encoders~\citep{khandelwal2022simple, majumdar2022zson} or object-grounding tools~\citep{gadre2023cows, dorbala2023can} for navigation.
\citet{yu2023l3mvn} utilizes an LLM as the planning backbone with a captioning model for perception. Following them, many improvements have been proposed~\citep{cai2024bridging, chen2023not, zhou2023esc, shah2023navigation}.
While these works still focus on a small set of target objects, we explore the potential of recent LVLMs for open-vocabulary object navigation.


\begin{figure*}[t]
    \centering
    \includegraphics[width=0.9\textwidth]{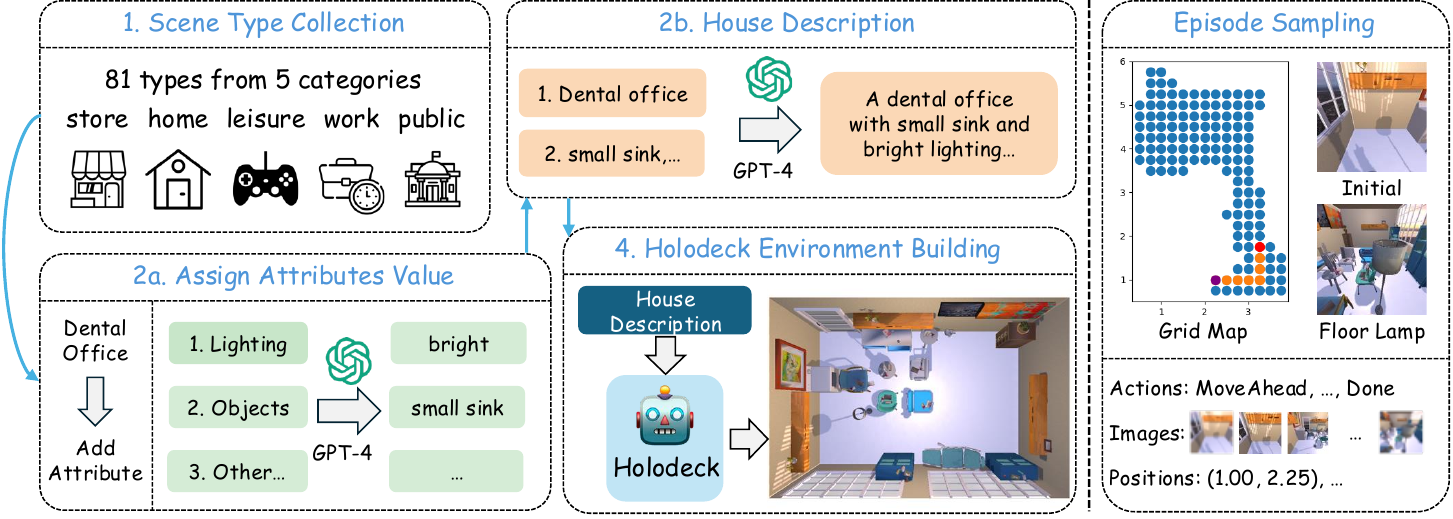}
    \caption{Data collection process. On the left, we show the process of collecting scenes. We prompt GPT-4o to collect textual house descriptions and use Holodeck to build houses. On the right, we show an episode built in the house, where we use BFS to find the shortest path. Then, actions and observations are collected.}
    \vspace{-0.30cm}
    \label{fig:method_figure}
\end{figure*}

\section{Task Definition}
\label{sec:task_definition}
We study the open-vocab object navigation task involving an agent in an environment to find the target object belonging to a given category. The object categories in the open-vocabulary setting are denoted by $C = \{c_0, c_1, c_2, \ldots\}$, which is not restricted to a predefined set. The houses can be described by $H=\{h_0, h_1, \ldots, h_n\}$, and $n$ is the total number of scenes. In each episode, the embodied agent is initialized at a random position $p_i$ with rotation $r_i$ in a house $h_i$. Then, the embodied agent is required to perform instance-level object navigation, where it gets a target object within the category $c_i$ at the position $o_i$. Thus, an episode can be represented as $E^{(i)} = \{h_i, p_i, r_i, c_i, o_i\}$. At each time step $t$, the embodied agent observes the environment and predicts the next action $a_t$. Following previous work~\citep{yu2023l3mvn, zhu2024navi2gaze}, the observation comprises an RGB image of the egocentric view and the agent status (i.e., its position and rotation). 

Our new houses are all collected on the AI2THOR platform~\citep{kolve2017ai2}, and thus the action space, signified as $A$, covers: \moveahead, \rotateright, \rotateleft, and \done. 
By default, the \moveahead action moves the agent 25 centimeters, and \rotateleft (or \rotateright) rotates the agent 90 degrees. 
The \done action is used by the agent to indicate that the navigation task is complete.
When the agent takes the \done action or reaches the max action limit, the episode is considered successful if the distance to the target object is below 1.5 meters. Based on the actions, an environment can be discretized as a 0.25 $\times$ 0.25-meter grid map of all reachable positions. 

\section{Data Collection for \scenedataname and \trajdataname}
Existing object navigation studies only focus on limited types of scenes and objects. To fill this gap, we first curate a large-scale scene dataset \scenedataname, featuring 4,614 scenes. Then, 23K shortest-path episodes with 5,707 kinds of target objects were sampled using breadth-first search (BFS), forming the \trajdataname dataset. We illustrate the details in \cref{fig:method_figure}.

\subsection{Scene Collection for \scenedataname}
We adopt Holodeck~\citep{yang2024holodeck} to build scenes, easing human labor. Holodeck takes textual house descriptions as input and uses GPT-4 to decide the layout, styles, and object selections. To collect diverse houses, we first manually compile 81 scene types across five categories by supplementing the MIT scene dataset~\citep{quattoni2009recognizing}, like music studio and home office.

\begin{figure}[t]
  \centering
  \includegraphics[width=0.8\columnwidth]{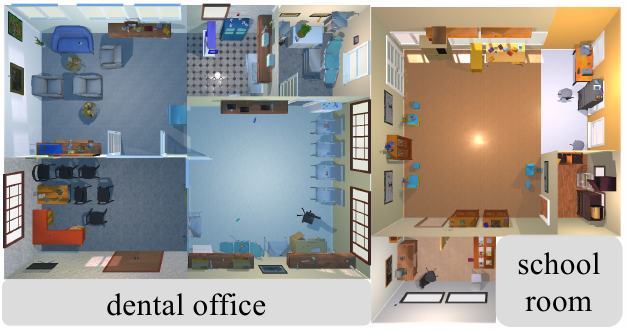}
  \caption{Examples of houses with different scene types.}
  \vspace{-0.45cm}
  \label{fig:scene_type_example}
\end{figure}

Then, we build textual house descriptions based on randomly chosen scene types by adding house attributes. We consider 12 house attributes, such as room style, users of the room, etc. We randomly sample 1-3 attributes and prompt GPT-4 to assign specific values to them. Given the scene type and attribute values, a house description is then written by GPT-4. Here, we prompt GPT-4 under the in-context learning setting~\citep{brown2020language}, and five exemplars are provided to generate the description for the final test example. See full details of scene types and attributes in \cref{app:scene_type} and concrete prompts in \cref{app:textual_description}.
With a strict data filtering (\cref{app:house_des_post}), we obtained 4,614 houses. We present a few houses in \cref{fig:scene_type_example} and more examples in \cref{app:more_house_example}. 

\subsection{Episode Collection for \trajdataname}
There are three steps in our episode sampling method. 
First, we sample the agent's initial position and the target object. Then, we use BFS to find the shortest paths in the 0.25 $\times$ 0.25-meter grid map. Finally, we obtain the action sequence and corresponding observations. We show an example on the right side of \cref{fig:method_figure}.

First, as discussed in \cref{sec:task_definition}, houses are discretized as grid maps of the fixed step size. We randomly sample an initial position from the grip map for the agent. Then, a target object is randomly sampled from all available objects in the environment. To encourage diversity, objects of the same type as those sampled in previous episodes are removed from the pool when sampling for new episodes in the same house. We also impose that the target objects are between 0.3m and 2.0m in height to ensure they are observable to the agent.

Second, we find the shortest path in the grid map that navigates a multi-room, cluttered environment. We can obtain the ground truth information from AI2-THOR, like reachable positions and objects' coordinates. We use BFS to find the shortest path from the initial position to the target object. More details are in \cref{app:planner_code}. This ground truth information is not provided to agents for inference and is only used to produce episodes for training. 

With the shortest path, we then derive the sequence of actions needed to achieve navigation. Basically, between two adjacent positions in the shortest path, we add a \moveahead action if the agent's rotation remains unaltered. Otherwise, we first use \rotateright or \rotateleft to adjust the orientation and then add a \moveahead action (more details in \cref{app:action_derive}). Thus, we obtain an action sequence that can steer the agent to the target object. We execute all actions in AI2THOR and collect observations at each step, including the agent's status and egocentric images.

\begin{table}[t]
\small
\setlength{\tabcolsep}{3.5pt}
\centering
\begin{tabular}{l|ccc}
\hline
\textbf{Metric}                & \textbf{ProcTHOR} & \textbf{iTHOR} & \textbf{DivScene} \\ \hline
Scene types  & 4 & 4  & 81  \\ 
Rooms per house & 3.78 & 1.00  & 2.35 \\ 
Room size (m\textsuperscript{2}) & 25.21 & 34.24          & 30.15 \\
Obj. types per house & 16.25 & 30.92 & 32.17 \\ 
Obj. per house & 35.64  & 47.26  & 111.42 \\ 
Obj. types & 38  & 116   & 22,696  \\ 
\hline
\end{tabular}
\caption{A comparison of ProcTHOR, iTHOR, and our dataset. ``Obj.'' refers to objects.}
\label{tab:dataset_comparison}
\vspace{-1.2em}
\end{table}

\subsection{Statistics and Comparison}
\label{sec:dataset_stat}
In \scenedataname, we collected 4,614 houses across 81 scene types. To the best of our knowledge, this dataset covers the widest range of scenes. We show the diversity of our collected houses in \cref{fig:top_room_type} by plotting most types under each category. Thanks to Objaverse~\citep{deitke2023objaverse}, the collected houses contain objects from 22,696 different types, including very common objects such as fridges, beds, shelves, and sofas, and rare objects such as multicolored bookshelf and vintage wooden bench. 

Then, we randomly pick one house of each scene type to build a test set of 81 houses. Similarly, we randomly select 27 houses of distinct scene types as the validation set. \scenedataname is divided into training, validation, and test sets, covering 4,506, 27, and 81 houses. 
On the training set, we sample five episodes in each scene with different target objects. 
Four episodes are selected in each house in the evaluation sets to balance the evaluation efficiency and accuracy. 
In total, the \trajdataname dataset contains 22,962 episodes and 5,707 different kinds of target objects. 

We provide a detailed comparison of scene complexity between our dataset and the other two datasets: iTHOR~\cite{weihs2021visual} and ProcTHOR~\cite{deitke2022️}. As shown in \cref{tab:dataset_comparison}, our dataset includes the most objects and scenes, and room sizes and numbers in our dataset remain consistent with manually created ones in iTHOR. More statistics are shown in \cref{app:data_diversity}.

\section{NatVLM}
In this section, we describe our \modelname model. It is fine-tuned from Idefics 2~\citep{laurenccon2024matters} through imitation learning on shortest paths from \trajdataname to enhance the navigation ability. At each time step $t$, the model processes environment observations $s_t$ and generates the next action $a_t$ in a manner consistent with instruction tuning. \cref{fig:intro_figure} provides an overview of \modelname.

\subsection{Instruction Compilation}
We manually build instructions for each time step in an episode, which contains four parts: 
\begin{enumerate*}[label=(\roman*)]
    \item the instruction provides a brief introduction to the object navigation task, such as possible actions and step length;
    \item we add the current environment observation, including the target object, its position, the agent's position and rotation, and visual observation;
    \item we provide the agent's positions and actions for the recent $M=8$ steps, along with the visual observations from the recent $K=4$ steps, to help the agent capture navigation history;
    \item the instruction asks the model to predict the next action by considering the current observation and navigation history in accordance with the CoT explanation in the responses.
\end{enumerate*}
We show the specific instruction template in \cref{app:instruction_template}.

\subsection{Response with CoT Explanation}
We tune the Idefics 2 to generate the next action when instructions are given. Specifically, we encode actions as natural language and use the model's generative capabilities to decode the next action. Nonetheless, we found that merely requiring LVLMs to output the next action leads to unsatisfactory results. The model only grasps the surface-level styles of the prompt but misses the underlying rationales of object navigation.

To enhance its understanding of the object navigation task, we manually build responses with CoT explanation traces during the navigation process. The structure of the responses covers three steps. In the first step, we have the agent compare its current position with the target and determine whether it needs to move forward or take other actions. After this, the agent is asked to check the obstacles in the visual observation to see whether it needs to rotate. In the last step, the agent gives the final decision based on the analyses in the first two steps. In contrast to writing explanations with LLMs~\citep{mitra2023orca}, we manually write the prompt template of explanation traces, leaving the position information to be filled in with coordinates at each step. Meanwhile, we employ a few postprocessing steps, like action balancing and conflict filtering. We show concrete prompts and postprocessing details in \cref{app:response_template,app:data_postprocessing}, respectively.

\subsection{Imitation Learning Objective}
With collected instructions and responses, we employ imitation learning to train \modelname to mimic decision-making from shortest-path demonstrations. Specifically, we adopt the \textit{Behavior Cloning} (BC) framework, where the goal is to learn a policy $\pi_\theta(a_t|s_t)$ that maps the current state $s_t$ (environment observation) to the action $a_t$ at each time step $t$. We minimize the negative log-likelihood of the shortest-path actions over collected observation-action pairs $(s_t, a_t)$ obtained from trajectories in \trajdataname, which is denoted as $\mathcal{D}$. The objective function is given by:

\vspace{-1.2em}
\begin{equation}
\mathcal{L}_{\text{BC}}(\theta) = - \mathbb{E}_{(s_t, a_t) \sim \mathcal{D}} \big[ \log \pi_\theta(a_t | s_t) \big],
\end{equation}
where $\theta$ represents the learnable parameters of Idefics 2. As we aforementioned, the state $s_t$ is included in the instruction, and the action $a_t$ includes the CoT explanation besides the next action.

\section{Experiment}
We conduct extensive experiments to evaluate both off-the-shelf LVLMs and LLMs, as well as models that have been fine-tuned using \trajdataname.  

\subsection{Metrics}
A model navigates in a house until it chooses the \done action or reaches the maximum action limit of 200 steps. We report the metrics Success Rate (SR), Success
weighted by Path Length~\citep[SPL;][]{anderson2018evaluation}, and Success
weighted by Episode Length~\citep[SEL;][]{eftekhar2023selective}. An episode is considered successful when the target object appears in the agent egocentric observation and is less than 1.5 meters away. Specifically, SR and SPL are computed as $\frac{1}{N}\sum_{i=1}^{N}S_i$ and $\frac{1}{N}\sum_{i=1}^{N}S_i\frac{l_i}{max(l_i,p_i)}$, where $N$ is the number of episodes, $S_i$ is the indicator of success, $l_i$ is the length of the shortest path, and $p_i$ is the length of the predicted trajectory. For SEL, we replace $l_i$ and $p_i$ with the action number of the shortest and predicted paths.

\begin{table*}[t]
    \small
    \setlength{\tabcolsep}{3.7pt}
	\centering
	\begin{tabular}{l|l||ccc|ccc|ccc}
	\toprule
        \multirow{2}{*}{\textbf{Methods}}&\multirow{2}{*}{\textbf{Backbone}}&\multicolumn{3}{c|}{\textbf{Valid}} &\multicolumn{3}{c|}{\textbf{Test}}&\multicolumn{3}{c}{\textbf{All}}\\
	&&\textbf{SR}&\textbf{SPL} &\textbf{SEL} &\textbf{SR}&\textbf{SPL} &\textbf{SEL} & \textbf{SR}&\textbf{SPL}&\textbf{SEL} \\
            \midrule
            \textbf{Random} & \multicolumn{1}{c||}{-} &9.26&8.19&9.26&6.79&5.77&6.79&8.03&6.98&8.03 \\
            \midrule
            \multirow{4}{*}{\textbf{Blind LLMs}}&Llama 2 (7B)&8.33&7.26&3.64&9.57&7.63&6.28&8.95&7.45&4.96\\
            &Llama 2 (13B)&9.26&7.69&3.12&10.19&8.62&4.14&9.72&8.15&3.63\\
            &Llama 3.1 (8B)&11.11&9.40&5.56&12.04&9.50&6.19&11.57&9.45&5.88\\
            &Mistral (7B)&8.33&7.16&4.13&9.88&7.89&3.78&9.11&7.53&3.96\\
            \midrule
            
            \multirow{4}{*}{\textbf{LLMs w/ Captions}} &Llama 2 (7B)&11.11&9.30&5.06&12.96&10.90&8.19&12.04&10.10&6.62\\
            &Llama 2 (13B)&9.26&7.56&4.03&12.35&9.93&5.95&10.80&8.74&4.99\\
            &Llama 3.1 (8B)&12.96&10.73&2.75&16.67&13.50&6.28&14.82&12.12&4.52\\
            &Mistral (7B)&11.11&9.65&3.43&11.76&9.65&2.72&11.43&9.65&3.07\\
            \midrule
            
            \multirow{7}{*}{\textbf{Open LVLMs}}&Qwen-VL (7B)&10.19&8.75&9.14&7.41&6.05&6.66&8.80&7.40&7.90\\
            &Llava 1.5 (7B)&12.04&10.07&9.88&12.35&10.03&10.30&12.20&10.05&10.09\\
            &Llava 1.5 (13B)&12.04&10.50&11.05&10.62&8.73&9.75&11.33&9.62&10.40\\
            &Idefics 2 (8B)&21.30&17.88&14.31&20.68&17.18&16.49&20.99&17.53&15.40\\
            & Qwen2.5-VL (7B) &11.11&8.96&10.94&10.20&8.14&9.70&10.65&8.55&10.32\\
            & InternVL3 (8B) &11.43&9.62&10.26&11.46&9.59&11.46&11.45&9.61&10.86\\
            & Gemma3 (12B) &15.74&12.96&5.94&22.22&18.36&5.17&18.98&15.66&5.55\\
            \midrule
            
            \multirow{2}{*}{\textbf{API LVLMs}}&GPT-4v&33.33&28.79&18.81&32.10&26.39&18.26&32.72&27.59&18.54\\
            &GPT-4o&37.04&31.82&29.47&38.27&31.74&27.92&37.66&31.78&28.70\\
		\midrule
        \textbf{Fine-Tuned LVLMs}&Idefics 2 (8B)&29.63&25.01&23.18&26.54&22.11&21.42&28.09&23.56&22.30\\
            \midrule
            \textbf{\modelname (Ours)} &Idefics 2 (8B)&\textbf{57.41}&\textbf{47.84}&\textbf{47.90}&\textbf{54.94}&\textbf{44.45}&\textbf{45.83}&\textbf{56.17}&\textbf{46.15}&\textbf{46.86} \\
		\bottomrule
	\end{tabular}
     \caption{Performance of all models \trajdataname. ``All'' means the average of both evaluation sets. The highest scores are bolded. In the ``LLMs w/Captions'' baseline, we use Llava 1.5 as the captioning model.}
    \label{tab:main_eval}
    \vspace{-1em}
\end{table*}

\subsection{Evaluated Methods}
Besides \modelname, we comprehensively evaluate four off-the-shelf methods and one fine-tuned method on our sampled dataset\footnote{We do not include complicated frameworks built on LVLMs as they are out of the evaluation scope. More in \cref{app:zero_shot_transfer}.}:

\noindent\textbf{Blind LLMs} are text-only LLMs that simply predict the next action based on the textual instruction without considering any visual information. This method references how far we can get solely using prior world knowledge and random guessing. For the LLM choice, we evaluate Llama 2 (7B, 13B)~\citep{touvron2023llama}, Llama 3.1 (8B)~\citep{dubey2024llama}, and Mistral (7B)~\citep{jiang2023mistral}.

\noindent\textbf{Socratic LLMs w/ Image Captions} is the simplest method that leverages visual information. Here, we use an image captioning model to convert egocentric images into language descriptions, allowing LLMs to obtain the content of visual information. We employ Llava 1.5~\citep{liu2024improved} as the captioning model while using the same LLMs as those in the ``Blind LLM'' baseline. 

\noindent\textbf{Open-Source LVLMs} are directly tested without any further tuning. They are capable of processing images in addition to textual queries. Here, we test Llava 1.5 (7B, 13B)~\citep{liu2024visual, liu2024improved}, Qwen-VL (7B)~\citep{bai2023qwen}, Idefics 2 (8B)~\citep{laurenccon2024matters}, Qwen2.5-VL (7B)~\citep{bai2025qwen2}, InternVL3 (8B)~\citep{zhu2025internvl3}, and Gemma3 (12B)~\citep{team2025gemma}.

\noindent\textbf{API-based LVLMs} we evaluated include GPT-4v~\citep{2023GPT4VisionSC} and GPT-4o~\citep{2024GPT4oSC}. They can process multiple images and achieve state-of-the-art performance on multimodal tasks.

\noindent\textbf{Fine-Tuned LVLM:} For fine-tuned LVLMs on our data, we also evaluated fine-tuning Idefics 2~\citep{laurenccon2024matters} on our sampled trajectories without any CoT explanation traces.
\begin{table*}[t]
    \small
    \setlength{\tabcolsep}{4.5pt}
	\centering
	\begin{tabular}{l||ccc|ccc|ccc}
	\toprule
        \multirow{2}{*}{\textbf{Methods}}&\multicolumn{3}{c|}{\textbf{Valid}} &\multicolumn{3}{c|}{\textbf{Test}}&\multicolumn{3}{c}{\textbf{Test Diff}}\\
	&\textbf{SR}&\textbf{SPL} &\textbf{SEL} &\textbf{SR}&\textbf{SPL} &\textbf{SEL} & \textbf{$\Delta_{SR}$} & \textbf{$\Delta_{SPL}$} & \textbf{$\Delta_{SEL}$} \\
            \midrule            \modelname (Ours)&\textbf{57.41}&\textbf{47.84}&\textbf{47.90}&\textbf{54.94}&\textbf{44.45}&45.83 & - &- &-\\
            \midrule
            $\diamond$ w/o ET&29.63&25.01&23.18&26.54&22.11&21.42&$\downarrow$28.40&$\downarrow$22.34&$\downarrow$24.41\\
            $\diamond$ w/o ET \& w Gold &28.70&24.12&23.38&30.86&25.46&25.58&$\downarrow$24.08&$\downarrow$18.99&$\downarrow$20.25\\
            \midrule
            $\diamond$ w Gold Label &59.26&49.01&51.33&62.96&50.54&54.12&$\uparrow$8.02&$\uparrow$6.09&$\uparrow$8.29\\
            $\diamond$ w Diff-EQ&54.63&45.02&46.48&54.32&43.59&\textbf{46.84}&$\downarrow$0.62&$\downarrow$0.86&$\uparrow$1.01\\
		\bottomrule
	\end{tabular}
         \caption{The ablation study. $\Delta_*$ columns show score differences. We remove explanation traces and test different methods of position comparisons. We bold the highest scores except for the gold label test ($\diamond$ w Gold Label).}
    \label{tab:ablation}
\end{table*}

\subsection{Main Evaluation}
We present the results of all off-the-shelf and fine-tuned methods on the validation and test sets in \cref{tab:main_eval}. We also include the performance of selecting a random action at each step (i.e., \textbf{Random}) as a reference. In general, the fine-tuned \modelname achieves the best performance on object navigation, exceeding other methods by a large margin. For example, \modelname can successfully navigate to 57.41\% of episodes on the validation set, increasing by about 20\% compared to the GPT-4o baseline. Meanwhile, according to the higher SPL and SEL on both test and validation sets, \modelname can navigate to target objects with better efficiency.

For the off-the-shelf methods, the blind LLMs achieve performance slightly higher than the random results. For example, Llama 3.1 (8B) achieves a success rate of 11.57\%, about 4 points higher than the random guess. In the meantime, we observe that the performance of all LLMs only improved marginally when we added the captioning model to provide additional perceptual information. This result shows that the captioning model can miss important image content details, leading to unsatisfactory improvement. On the other hand, we find that LVLMs can achieve the best results across all baselines. For example, the closed-source LVLMs, GPT-4v and GPT-4o, can successfully navigate to target objects in more than 30\% cases. In addition, Idefics 2 (8B) attains success rates exceeding 20\% on both validation and test sets.

\subsection{Ablation Evaluation}
To better understand the role of CoT explanation traces in \modelname, we conduct two ablation studies to analyze its contribution. First, we verify the efficacy of CoT explanation traces. Then, we analyze different ways to compare the positions of the agent and target object. The results of experiments are shown in \cref{tab:ablation}. 

First, we remove explanation traces in the instruction tuning data ($\diamond$ w/o ET). Thus, we fine-tune the agent to only generate the next action. We find that the performance of our agent drastically drops, verifying that explanation traces can help LVLMs to better understand the underlying rationales of object navigation. Furthermore, we enhance the agent's input by providing the gold label of the positional difference between the agent and the target object ($\diamond$ w/o ET \& w Gold). We observe that the gold labels cannot help much, as the performance only fluctuates somewhat.  

Then, we study the effects of the position comparison, the crucial component in our explanation traces. The default prompt is to directly generate the position difference: ``the difference to the target object is \textbf{[position\_diff]}'' as shown in \cref{tab:response_template_move_done}, where \textbf{[position\_diff]} is a placeholder. First, we provide the global label of positional differences in the input instructions ($\diamond$ w Gold Label). The results in \cref{tab:ablation} show that providing global labels can improve the performance of our agent, suggesting that our agent may occasionally compute the positional difference with errors. Next, we test the difference-equation prompt ($\diamond$ w Diff-EQ), where we fine-tune our agent to generate an equation for computing the positional difference. However, writing the equation of computation only leads to small variations in performance. All the abovementioned prompts are shown in \cref{app:ablation_stduy}.

\begin{figure}[t]
	\centering
        \includegraphics[width=\columnwidth]{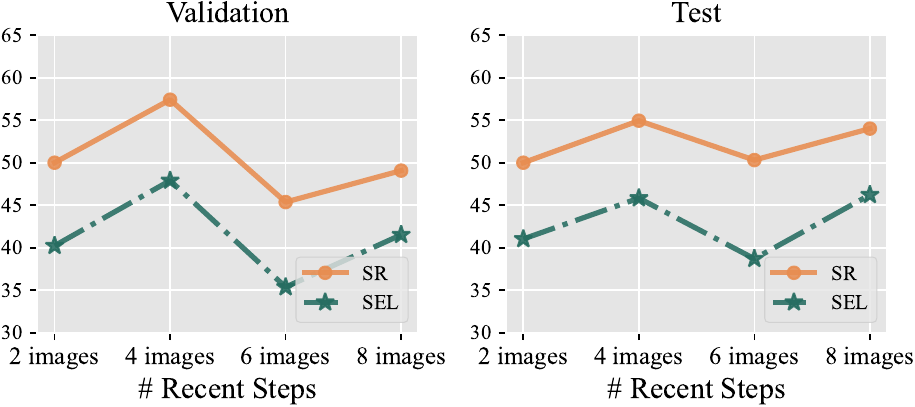}
	\caption{Design investigation. We provide different numbers of recent visual observations with the embodied agent. See scores of all metrics in \cref{app:image_number}} 
    \label{fig:image_number}
\end{figure}

\subsection{Design Investigation}
In this experiment, we thoroughly investigate a few design decisions regarding the image number and step number of recent positions and actions. We also test \modelname with different prompts shown in \cref{app:image_number,app:action_number}

The default design provides \modelname with four images. In addition, we test its performance using different numbers of input images, including 2, 6, and 8 images. The results are plotted in \cref{fig:image_number}. First, we observe a decline in the agent's performance when provided with only two images, showing that using fewer images leads to worse performance. For instance, our agent finishes the navigation successfully only 50\% of the time, underperforming the 4-image baseline. Moreover, increasing the number of images to 6 or 8 does not result in further improvements. Thus, we choose to provide \modelname with four images for the tradeoff between accuracy and efficiency.

We also investigate the effect of recent positions and actions on navigation performance. By default, we provide \modelname with information about the recent 8 steps. Here, we test the performance when we provide the positions and actions of 4, 12, and 16 steps. As illustrated in \cref{tab:action_number}, a substantial performance improvement is evident when increasing the number of steps from 4 to 8. However, further increases in step count yield limited returns. Thus, we provide our agent with 8 steps.

\begin{table}[t]
\centering
\small
\setlength{\tabcolsep}{4.5pt}
\begin{tabular}{c||ccc|ccc}
	\toprule
        \multirow{2}{*}{\textbf{Number}}&\multicolumn{3}{c|}{\textbf{Valid}} &\multicolumn{3}{c}{\textbf{Test}}\\
	&\textbf{SR}&\textbf{SPL} &\textbf{SEL} &\textbf{SR}&\textbf{SPL} &\textbf{SEL}\\
            \midrule
4 steps&46.30&38.66&39.82&45.99&37.45&39.60\\
8 steps&\textbf{57.41}&\textbf{47.84}&\textbf{47.90}&\textbf{54.94}&\textbf{44.45}&\textbf{45.83}\\
12 steps&42.59&35.92&36.43&50.93&41.15&43.20\\
16 steps&51.85&43.01&42.72&53.40&43.03&44.08\\
		\bottomrule
\end{tabular}
\caption{Our agents receive various numbers of recent actions and positions. See
all metrics in \cref{app:action_number}.}
\label{tab:action_number}
\end{table}

\begin{table}[t]
\centering
\small
\setlength{\tabcolsep}{3.5pt}
\begin{tabular}{c||ccc|ccc}
	\toprule
        \multirow{2}{*}{\textbf{\% Data}}&\multicolumn{3}{c|}{\textbf{Test}}&\multicolumn{3}{c}{\textbf{Test Diff w GPT-4o}}\\
	&\textbf{SR}&\textbf{SPL} &\textbf{SEL} & \textbf{$\Delta_{SR}$} & \textbf{$\Delta_{SPL}$} & \textbf{$\Delta_{SEL}$}  \\
            \midrule
20&38.89&31.40&32.13&$\uparrow$0.62&$\downarrow$0.34&$\uparrow$4.21\\
40&38.27&31.00&30.79&$\downarrow$0.00&$\downarrow$0.74&$\uparrow$2.87\\
60&52.12&42.25&44.40&$\uparrow$13.85&$\uparrow$10.51&$\uparrow$16.48\\
80&49.69&40.26&42.49&$\uparrow$11.42&$\uparrow$8.52&$\uparrow$14.57\\
100&\textbf{54.94}&\textbf{44.45}&\textbf{45.83}&$\uparrow$16.67&$\uparrow$12.71&$\uparrow$17.91\\
\bottomrule
\end{tabular}
\caption{Few-shot learning ability. See scores of the validation set and more prompts in \cref{app:few_shot}.}
\vspace{-0.4cm}
\label{tab:few_shot_experiment}
\end{table}

\subsection{Few-shot Learning Evaluation}
\modelname undergoes instruction tuning with an extensive training dataset, \trajdataname. In this section, we design a few-shot experiment to confirm its ability to generalize with fewer data. While the original training data contains five episodes per house, we train the model with only 1, 2, 3, and 4 episodes, representing 20\%, 40\%, 60\%, and 80\% of the full data. The results are shown in \cref{tab:few_shot_experiment}. For clarity, we also compare our models with the GPT-4o baseline. We can find that our agent has strong few-shot learning abilities. With only 20\% percent training data, our agent can perform similarly to GPT-4o and generalize well on unseen houses. Meanwhile, the results demonstrate a gradual performance improvement as we incrementally increase the data volume, with performance plateauing at approximately 80\% of the full dataset. We provide more analysis in \cref{app:few_shot}.

\begin{table}[t]
    \small
    \setlength{\tabcolsep}{3pt}
	\centering
	\begin{tabular}{l||ccc|ccc}
	\toprule
        \multirow{2}{*}{\textbf{Models}}&\multicolumn{3}{c|}{\textbf{iTHOR}} &\multicolumn{3}{c}{\textbf{ProcTHOR}}\\
	&\textbf{SR}&\textbf{SPL} &\textbf{SEL} &\textbf{SR}&\textbf{SPL} &\textbf{SEL} \\
            \midrule
            Qwen-VL (7B) &23.67&19.96&19.27 &10.83&9.04&6.28 \\
            Llava 1.5 (7B) &24.12&20.32&20.34 &16.04&13.53&14.02 \\
            Llava 1.5 (13B) &19.47&16.21&17.48 &13.12&11.07&11.85 \\
            Idefics 2 (8B) &28.54&23.39&18.49 &17.29&14.33&11.05 \\
                        \midrule
            \modelname &\textbf{72.79}&\textbf{59.34}&59.28 &53.12&44.37&43.04 \\
            $\diamond$ w/o ET &38.27&32.15&31.43 &31.25&26.87&26.20 \\
            $\diamond$ w Diff-EQ &72.32&58.98&\textbf{62.35} &\textbf{54.59}&\textbf{45.53}&\textbf{46.80} \\
		\bottomrule
	\end{tabular}
    \caption{Zero-shot transferring on iTHOR and ProcTHOR.}
    \label{tab:zero_transfer}
\end{table}

\subsection{Zero-shot Transferring Evaluation}
We further evaluate \modelname on other house datasets: iTHOR~\citep{weihs2021visual}, ProcTHOR~\citep{deitke2022️}, and HM3D~\cite{ramakrishnan2021habitat}. We directly use \modelname tuned on \trajdataname to conduct zero-shot transferring evaluation. Since we do not tune hyperparameters on these datasets, we treat each dataset as a whole test set without any validation set. We show the results on iTHOR and ProcTHOR in \cref{tab:zero_transfer}. We also report the performance of open-source VLMs and some ablated models as baselines. The results show that our agent surpasses all the baselines on both datasets, indicating that our framework has a strong ability to generalize in other environments. We include results on HM3D and setup details in \cref{app:zero_shot_transfer}.

\begin{figure}[t]
  \centering
  \includegraphics[width=\columnwidth]{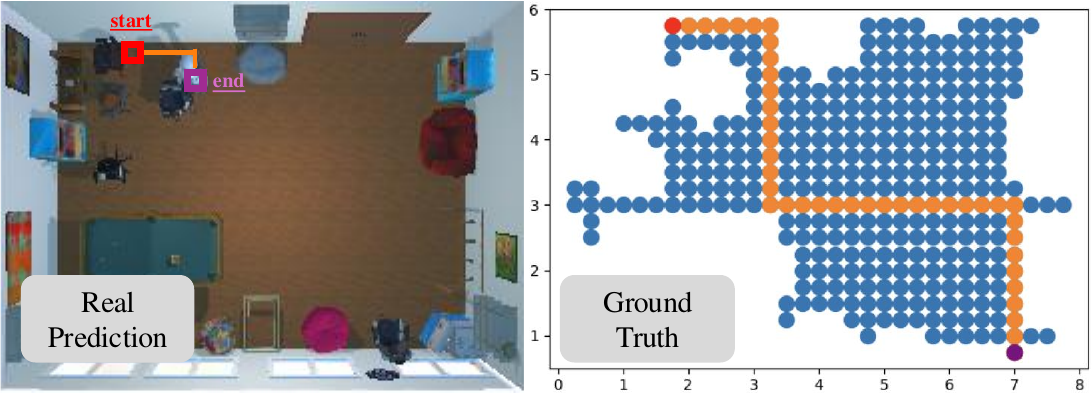}
  \caption{Error analysis. The agent needs to find a soda can in the lower right corner of a game room. The left image shows the predicted path, and the right image shows the ground truth in the grid map. The agent fails to find the soda can and wanders within a limited area.}
  \vspace{-0.1cm}
  \label{fig:case_study_example}
\end{figure}

\subsection{Case Study}
\label{sec:case_study}
In \cref{fig:case_study_example}, we provide an example of error analysis of \modelname. The left image shows the predicted path in the real scene, and the right image shows the ground truth in the corresponding grid map. Here, the agent needs to find a soda can after walking through the whole room. There are 46 actions in the shortest path. Instead of heading towards the goal, the agent just meanders in a limited area. This shows that \modelname cannot finish the navigation with a long trajectory and has limited exploration. We leave the development of models better at exploring environments for future work.

\section{Conclusion}
In this work, we first study the open-vocab object navigation and evaluate extensive LVLMs and LLMs. Meanwhile, we collect a large-scale scene dataset, \scenedataname, featuring 4,614 scenes. Over 22K episodes are sampled with the shortest paths by BFS. Our evaluation shows that current LVLMs and LLMs still fall short of the ability of open-vocab object navigation. Then, we also fine-tuned LVLMs with the shortest paths through imitation learning, where we also introduced CoT explanations. Experiments show that the navigation ability of LVLMs can be improved significantly. For future
work, a promising direction is to enable LVLMs to navigate over longer horizons.

\section*{Limitation}
While we conducted extensive experiments, our model still has some failure cases due to long navigation horizon or exploration, as we discussed in \cref{sec:case_study}.
This might be because, due to the limited context window, we can only provide existing models with recent historical information.  
Thus, a promising direction for future work is to improve the memory~\citep{du2025rethinking} or long-context~\citep{team2025gemma,wang2025mmlongbench} capabilities of LVLMs to enable navigation over longer horizons. 

\section*{Ethics Statement}
Our scene dataset \scenedataname is built upon the publicly available AI2THOR platform~\citep{kolve2017ai2} and the Holodeck framework~\citep{yang2024holodeck}. We further extend our experiments on iTHOR~\citep{weihs2021visual} and ProcTHOR~\citep{deitke2022️}, both of which are open-source datasets.
We provide the complete details of the implementation of our \modelname agent and baselines in \cref{app:implemntation}, including the learning rate, batch size, hard device, API access, etc. Meanwhile, we show the details of all the post-processing data in \cref{app:data_postprocessing,app:house_des_post}.

\section*{Acknowledgements}
The authors of this paper were supported by the ITSP Platform Research Project (ITS/189/23FP) from ITC of Hong Kong, SAR, China, and the AoE (AoE/E-601/24-N), the RIF (R6021-20) and the GRF (16205322) from RGC of Hong Kong, SAR, China. This work was done when Zhaowei Wang worked as an intern at Tencent AI Lab.

\newpage
\bibliography{custom}

\newpage
\appendix

\section{\scenedataname and \trajdataname Details}
\subsection{Scene Type Details and Attribute List}
\label{app:scene_type}
This section lists all scene types we collected by complementing the MIT scene dataset. In total, there are 81 scene types across five different categories, as shown in \cref{tab:scene_type}.

To collect diverse house descriptions, we add various attributes to a randomly sampled scene type. Here, we consider 12 different attributes as shown in \cref{tab:attribute_list}. We ask GPT-4 to assign a value to each attribute and write a house description.

\begin{table*}[t!]
    \small
	\centering
	\begin{tabular}{p{0.14\textwidth}||p{0.8\textwidth}}
	\toprule
        \textbf{Category}&\textbf{Scene Type} \\
            \midrule
            \textbf{Store} \newline (16 types) & bakery, grocery store, clothing store, deli, laundromat, jewellery shop, bookstore, video store, florist shop, shoe shop, toy store, furniture store, electronics store, craft store, music store, sporting goods store \\
            \midrule
            \textbf{Home} \newline (21 types) & bedroom, nursery, closet, pantry, children room, lobby, dining room, corridor, living room, bathroom, kitchen, wine cellar, garage, sunroom, cabinet, study room, apartment, home office, basement, attic, laundry room\\
            \midrule
            \textbf{Public spaces} \newline (9 types) & prison cell, library, waiting room, museum, locker room, town hall, community center, convention center, recreation center \\
            \midrule
            \textbf{Leisure} \newline (14 types) & buffet, fast-food restaurant, restaurant, bar, game room, casino, gym, hair salon, arcade, spa, concert hall, ski lodge, lounge, club \\
            \midrule
            \textbf{Working place} \newline (21 types) & hospital room, kindergarten, restaurant kitchen, art studio, classroom, laboratory, music studio, operating room, office, computer room, warehouse, greenhouse, dental office, TV studio, meeting room, school room, conference room, factory floor, call center, reception area, nursing station \\
		\bottomrule
	\end{tabular}
\caption{The categories and scene types used in our \scenedataname dataset. In total, there are five categories and 81 scene types in our dataset.}
    \label{tab:scene_type}
\end{table*}

\begin{table*}[t]
    \small
	\centering
	\begin{tabular}{l|l||l|l}
	\toprule
        \textbf{Attribute}&\textbf{Example Value} & \textbf{Attribute}&\textbf{Example Value} \\
        \midrule
            Room Style & victorian, rustic& Flooring & soft and cushioned, hard \\
            Objects in the Room &computers, desks, chairs, servers& Theme & industrial, contemporary \\
            Number of Rooms &single room& Lighting & bright, warm ambient \\
            Configurations & individual cubicles& Window & small, slightly slanted \\ 
            Users of the Room & children of various ages& Room Size & spacious, medium-sized \\
            Era &contemporary, modern & Wall Treatment & artistic paintings, calming color \\
            \midrule
		\bottomrule
	\end{tabular}
\caption{The 12 attributes we used to collect house descriptions. We also provide an example value of each attribute.}
    \label{tab:attribute_list}
\end{table*}

\subsection{Textual House Description Prompt}
\label{app:textual_description}
We use in-context learning to prompt the GPT-4 to write textual house descriptions of 81 different scene types. Here, we show the concrete prompt we used in \cref{tab:house_des_prompt}. We randomly sample 1-3 house attributes and ask GPT-4 to assign a value to them. Then, a house description is written based on the given scene type and attribute values. We use five exemplars in the in-context learning setting.

\begin{table*}[t]
\centering
\small
\begin{tabular}{p{0.96\textwidth}}
	\toprule
	\textbf{Task Instruction:} Create a detailed and fluent description for a house based on the given scene type and features in two steps. Step 1: provide the value of each feature. Step 2: write a short phrase to describe the scene type with the values. \\
	\midrule
	\textbf{Exemplar1 Input:} The given house type is ``arcade.'' The feature list is: ``(1) Objects in the room.'' \\
	\textbf{Exemplar1 Output:} Step 1: (1) a pool table\textbackslash n Step 2: An arcade with a pool table \\
    \midrule
    \textbf{Following Exemplars:} Exemplar 2, ..., Exmplar 5 \\
    \midrule
    \textbf{Testing Input:} The given house type is ``office.'' The feature list is: ``(1) Number of Rooms (2) Users of the Room (3) Configurations.''\\
    \bottomrule
	\end{tabular}
\caption{The prompt we used to collect textual house descriptions using GPT-4. Here, we use 5 exemplars in the in-context learning. We show one example here for saving space.}
\label{tab:house_des_prompt}
\end{table*}

\subsection{Postprocessing of Textual House Description}
\label{app:house_des_post}
After we collect textual house descriptions from GPT-4, we also introduce three filters to ensure their diversity and quality. 

\begin{itemize}[left=5pt]
\item We first introduce a ROUGE-L filter.
To encourage diversity, a new description is discarded when its ROUGE-L similarity~\citep{lin2004rouge} with any existing description is above 0.8, following previous works~\citep{wang2022self, wang2024absinstruct}. 

\item Second, if we cannot correctly parse the output in the first step and find the values of the given attributes, we remove the example. 

\item Third, if we cannot find the house description in the second step in the output from GPT-4, we remove the example. The last two steps mean that the output does not follow the output format specified in the instructions and exemplars (see \cref{tab:house_des_prompt}).

\end{itemize}

\subsection{Breadth-First Search for Shortest Path}
\label{app:planner_code}
We use a BFS-based planner to find the shortest path from the initial position to a target point on the grid map. Notice that the target object is not necessarily anchored to a point on the grid map for realism. Thus, we find the grid point nearest to the target object as the destination of the navigation. 

The algorithm is shown in \cref{alg:bfs_shortest_path}, which is based on a priority queue. We design the BFS-based planner to pick the path with the fewest rotations to make it easier for LLMs to imitate. In detail, we add more costs when rotation changes since the agent needs one more rotation action before moving ahead, as shown at the 12th line in \cref{alg:bfs_shortest_path}.

\subsection{Action Derivation Algorithm}
\label{app:action_derive}
We show the action derivation algorithm in \cref{alg:get_action_path}. Between two adjacent positions in the shortest path, we add a \moveahead action if the agent's rotation remains unaltered. Otherwise, we first use \rotateright or \rotateleft to adjust the orientation and then add a \moveahead action. After reaching the target object, we then rotate the agent so that the object is approximately centered in the agent's egocentric view.

\subsection{Data Diversity}
\label{app:data_diversity}
\begin{figure*}[t]
\begin{minipage}{0.48\textwidth}
\centering
\includegraphics[width=\linewidth]{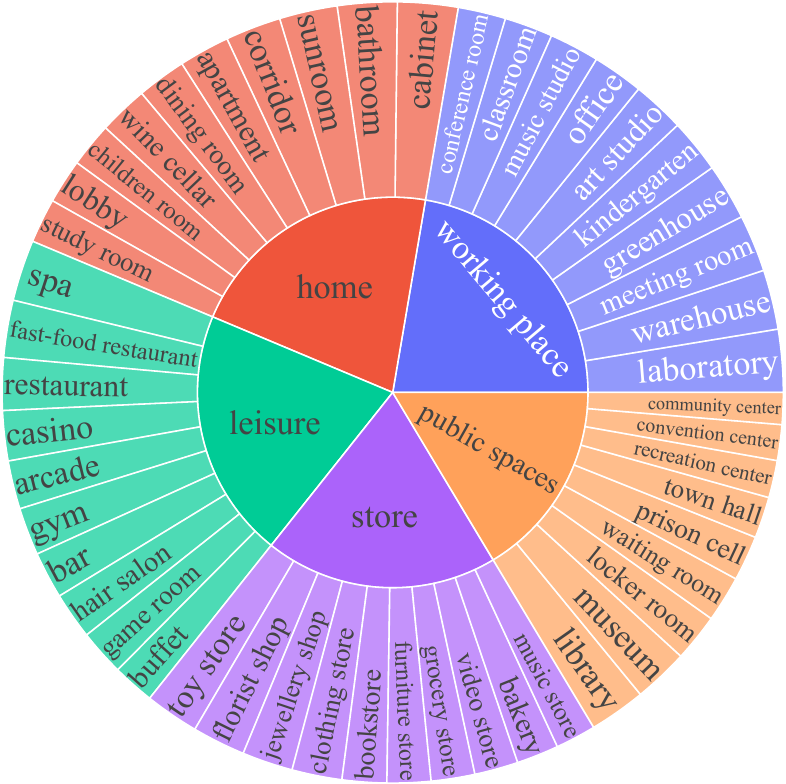}
\caption{The top 10 most common room types (outer circle) under each room category (inner circle) in the collected houses.}
\label{fig:top_room_type}

\end{minipage}\hfill
\begin{minipage}{0.48\textwidth}
\centering
\includegraphics[width=\linewidth]{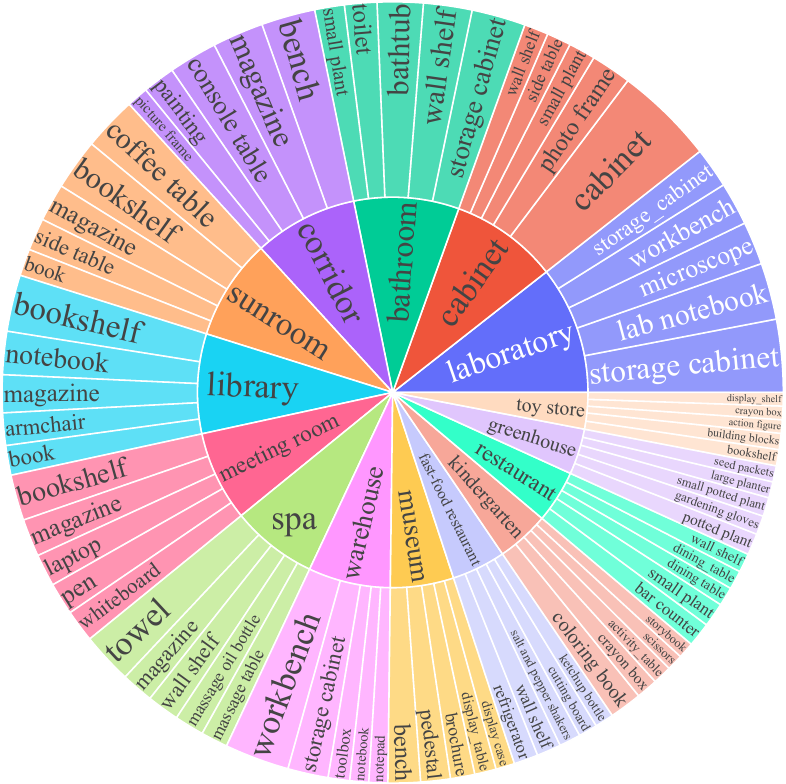}
\caption{The top 15 most common scene types (inner cycle) and their top 5 target object types (outer cycles).}
\label{fig:top_object_type}
\end{minipage}
\end{figure*}

To study what types of scenes are gathered under each category, we identify the category-type structure of houses in \scenedataname. We plot the top 10 most common scene types under each category in \cref{fig:top_room_type}. Then, we study the diversity of target objects in the episodes we sampled in \trajdataname. We plot the top 15 most common scene types and their top 5 target object types in \cref{fig:top_object_type}. Overall, we see quite diverse scenes and target objects in our datasets.

\begin{figure*}[t]
\centering
\begin{minipage}{0.95\textwidth}
\begin{algorithm}[H]
\caption{BFS Search for Shortest Paths}
\label{alg:bfs_shortest_path}
\begin{algorithmic}[1]
   \State {\bfseries Input:} $reachable\_pos, start\_point, end\_point, start\_rotation$
   \State Initialize priority queue $Q$, distance map $d\_map$, and parent map $p\_map$
   \State Enqueue $(0, start\_point, start\_rotation)$ to $Q$
   \While{$Q$ not empty}
       \State $(cost, current, rotation) \gets$ Dequeue from $Q$
       \If{$current = end\_point$}
           \State \Return ReconstructPath($p\_map$, $current$)
       \EndIf
       \For{$r \in \{$North, East, South, West$\}$}
           \State $neighbor \gets$ GetNeighbor($current, r$) \Comment{Four cardinal directions} 
           \If{$neighbor \in reachable\_pos$} 
               \State $newCost \gets cost +$ (1 if $r = rotation$ else 2) \Comment{More cost if rotation changed}
               \If{$neighbor$ not visited or $newCost < d\_map[neighbor]$}
                   \State $d\_map[neighbor] = newCost$ \Comment{Update distance}
                   \State $p\_map[neighbor] = current$ \Comment{Update parent}
                   \State Enqueue $(newCost, neighbor, r)$ to $Q$
               \EndIf
           \EndIf
       \EndFor
   \EndWhile
   \State \textbf{return} No path found
\end{algorithmic}
\end{algorithm}
\end{minipage}
\end{figure*}

\begin{figure*}[t]
\centering
\begin{minipage}{0.95\textwidth}
\begin{algorithm}[H]
\caption{Get a Sequence of Actions from a Shortest Path}
\label{alg:get_action_path}
\begin{algorithmic}[1]
\State {\bfseries Input:} $shortest\_path, start\_rotation, target\_object$
\State $agent\_rotation \leftarrow start\_rotation$
\State Initialize empty $action\_list$
\State $prior\_p \leftarrow shortest\_path[0]$
\For{$current\_p$ in $shortest\_path[1:]$}
    \State $path\_rotation \leftarrow$ compute\_path\_rotation($current\_p$, $prior\_p$) 
    \If{$path\_rotation \neq agent\_rotation$}
        \State Append appropriate rotation action(s) to $action\_list$ \Comment{RotateRight or RotateLeft}
        \State $agent\_rotation \leftarrow path\_rotation$
    \EndIf
    \State Append ``MoveAhead'' to $action\_list$
\EndFor
\State $object\_rotation \leftarrow$ compute\_object\_rotation($target\_object, shortest\_path[-1]$)
\If{$object\_rotation \neq agent\_rotation$}
    \State Append final rotation action(s) to $action\_list$ \Comment{Adjust the view for the target object}
\EndIf
\State Append ``Done'' to $action\_list$
\State \textbf{return} $action\_list$
\end{algorithmic}
\end{algorithm}
\end{minipage}
\end{figure*}

\section{\modelname Instruction Data}
In this section, we give the concrete prompts used in fine-tuning \modelname.
\subsection{Instruction Template}
\label{app:instruction_template}
We show the instruction template we used in the \cref{tab:instruction_template}. There are four parts in the instruction template, including a brief task introduction, episode-specific information, the status of recent steps, and the prediction steps that need to be considered. We leave a lot of placeholders for the episode and step information.

\begin{table*}[t!]
\centering
\small
\begin{tabular}{p{0.96\textwidth}}
    \toprule
    \textbf{1. Brief Introduction:} You are an agent placed in a 3D environment. Your step length is 0.25 meters, and your rotation degree is 90. \newline
    The possible actions are: \newline
    1. MoveAhead: Moves the agent forward by 0.25 meters in the direction it is currently facing.
    For example, if the agent is at (x, y) facing 0 degrees (north), MoveAhead will result in (x, y + 0.25). If the agent is facing 90 degrees (east), MoveAhead will result in (x + 0.25, y). If the agent is facing 180 degrees (south), MoveAhead will result in (x, y - 0.25). If the agent is facing 270 degrees (west), MoveAhead will result in (x - 0.25, y). \newline
    2. RotateRight: Rotate right for 90 degrees (clockwise). \newline
    3. RotateLeft: Rotate left for 90 degrees. (counterclockwise). \newline
    4. Done: Indicate that you are near to the target object and finish the task. 
    \\ \midrule
    \textbf{2. Episode-Specific Information:} You need to find a \textbf{[obj\_type]} at the position \textbf{[obj\_pos]}. To achieve this, we recommend you move to the position \textbf{[grid\_obj\_pos]} with a rotation of \textbf{[grid\_obj\_rotation]}. \newline 
    Currently, you are at \textbf{[agent\_pos]} with a rotation of \textbf{[agent\_rotation]}. 
    \\ \midrule
    \textbf{3. Status of Recent Steps:} The history of recent states are: \newline
    Position: \textbf{[recent\_agent\_pos]}, Rotation: \textbf{[recent\_agent\_rotation]}, Action: \textbf{[recent\_action]} \newline
    \ldots \newline
    Position: \textbf{[recent\_agent\_pos]}, Rotation: \textbf{[recent\_agent\_rotation]}, Current View: \textbf{[recent\_agent\_image]}, Action: \textbf{[recent\_action]}
    \\ \midrule
    \textbf{4. Prediction Steps:} Please generate the next step given the above states with the following steps: 1) Consider your rotation and position. 2) Check the images to see obstacles or the target object. 3) Decide the action. \\
    \bottomrule
	\end{tabular}
\caption{Instruction Template we used to fine-tune the LVLM: Idefics 2. There are four steps in the template, and we leave step-wise and episode information with placeholders. \textbf{[obj\_type]} and \textbf{[obj\_pos]} are the type of target object and its location. On the grid map, we also provide the nearest point on the grid map \textbf{[grid\_obj\_pos]} with a rotation \textbf{[grid\_obj\_rotation]}. \textbf{[agent\_pos]} and \textbf{[agent\_rotation]} are the agent's position and rotation. There are also placeholders for the recent status: \textbf{[recent\_agent\_pos]}, \textbf{[recent\_agent\_rotation]}, \textbf{[recent\_agent\_image]}, and \textbf{[recent\_action]}. Notice that we provide the information of the recent 8 steps and only provide the recent 4 images for inference efficiency.}
\label{tab:instruction_template}
\end{table*}

\subsection{Response Template}
\label{app:response_template}
Previous works usually collect explanation traces using GPT-4~\citep{mitra2023orca, mukherjee2023orca, ho2023large}. In contrast, we collect explanation traces with manually written templates. The templates and examples are shown in \cref{tab:response_template_move_done,tab:response_template_rotate}. For \rotateright and \rotateleft, we identified the three most common scenarios for rotation based on heuristic rules. Then, we wrote the template for each of them. The first scenario is that the distance difference between the agent and the target object becomes zero in the agent's rotation. Thus, the agent needs to navigate in the other direction. The second scenario involves the presence of obstacles in the agent's current path, necessitating a rotation to navigate around them. The final scenario involves adjusting the agent's rotation to center the target within its field of view, occurring at the end of the navigation process. 

\begin{table*}[t!]
\centering
\small
\begin{tabular}{p{0.96\textwidth}}
    \toprule

\multicolumn{1}{c}{\textbf{\moveahead}} \\ \midrule
Template: \newline
1) In the direction of my rotation, \textbf{[agent\_rotation]} degrees (\textbf{[cardinal\_direction]}), the difference to the target object is \textbf{[position\_diff]} m. I need to move further \textbf{[cardinal\_direction]}. \newline
2) There is no obstacle in front of me in recent images. \newline
3) MoveAhead \\ \midrule
Example: \newline
1) In the direction of my rotation, 90 degrees (east), the difference to the target object is 0.5m. I need to move further east. \newline
2) There is no obstacle in front of me in recent images. \newline
3) MoveAhead \\ \bottomrule \toprule
\multicolumn{1}{c}{\textbf{\done}} \\ \midrule
Template: \newline
1) My position and rotation are equal to the recommended one. \newline
2) I can see the target \textbf{[obj\_type]} in the image of the current state. \newline
3) Done
\\ \midrule
Example: \newline
1) My position and rotation are equal to the recommended one. \newline
2) I can see the target label marker in the image of the current state. \newline
3) Done \\
\bottomrule
\end{tabular}
\caption{Response templates we used to build CoT explanation traces for \moveahead and \done.}
\label{tab:response_template_move_done}
\end{table*}

\begin{table*}[h!]
    \small
	\centering
	\begin{tabular}{l||cc|cc}
	\toprule
        \multirow{2}{*}{\textbf{Action}}&\multicolumn{2}{c|}{\textbf{w/o Postproc}} &\multicolumn{2}{c}{\textbf{w/ Postproc}} \\
	&\textbf{\# Num}&\textbf{\% Prop} &\textbf{\# Num} &\textbf{\% Prop} \\
 \midrule
            \moveahead &221,598&77.94\% &57,760&49.32\% \\
            \rotateleft &19,596&6.89\% &18,412&15.72\% \\
            \rotateright &19,527&6.87\% & 18,403&15.72\% \\
            \done &23,610&8.30\% & 22,529&19.24\% \\
		\bottomrule
	\end{tabular}
\caption{The distribution of collected actions before and after post-processing. The original dataset is very imbalanced since most of the actions are \moveahead. Then, we downsample the \moveahead actions with a rate of 0.25. We also filtered the conflicting data.}
    \label{tab:data_postprocessing}
\end{table*}

\begin{table*}[t!]
\centering
\small
\begin{tabular}{p{0.96\textwidth}}
    \toprule
\multicolumn{1}{c}{\textbf{First scenario: distance difference becomes zero}} \\ \midrule
Template: \newline
1) In the direction of my rotation, \textbf{[agent\_rotation]} degrees (\textbf{[cardinal\_direction]}), the difference to the recommended position is 0.00m. Thus, I need to move in another direction, where the difference is \textbf{[other\_position\_diff]} m, and the rotation is \textbf{[other\_agent\_rotation]} degrees. \newline
2) Obstacles don't affect rotation. \newline
3) RotateRight/RotateLeft \\
\midrule
Example: \newline
1) In the direction of my rotation, 180 degrees (south), the difference to the recommended position is 0.00m. Thus, I need to move in another direction, where the difference is 1.25m, and the rotation is 90 degrees. \newline
2) Obstacles don't affect rotation. \newline
3) RotateLeft\\
\bottomrule
\toprule
\multicolumn{1}{c}{\textbf{Second scenario: Obstacles}} \\ \midrule
Template: \newline
1) In the direction of my rotation, \textbf{[agent\_rotation]} degrees (\textbf{[cardinal\_direction]}), the difference compared to the target object is \textbf{[position\_diff]} m. \newline
2) There are obstacles in front of me, as shown in current images. I need to rotate in another direction. In the other direction, the difference is \textbf{[other\_position\_diff]} m, and the rotation is \textbf{[other\_agent\_rotation]} degrees. \newline
3) RotateRight/RotateLeft \\ \midrule
Example: \newline 
1) In the direction of my rotation, 180 degrees (south), the difference compared to the target object is 1.50m. \newline
2) There are obstacles in front of me, as shown in current images. I need to rotate in another direction. In the other direction, the difference is 1.25m, and the rotation is 270 degrees. \newline
3) RotateRight
\\
\bottomrule
\toprule
\multicolumn{1}{c}{\textbf{Third scenario: View Adjustion}} \\ \midrule
Template: \newline
1) My position is the same as the recommended one: \textbf{[grid\_obj\_pos]}. However, my rotation is \textbf{[agent\_rotation]} degrees, facing \textbf{[cardinal\_direction]}. I need to adjust the rotation to center the target within its field of view. \newline
2) Obstacles don't affect rotation. \newline
3) RotateRight/RotateLeft \\ \midrule
Example: \newline
1) My position is the same as the recommended one: (0.50, 1.25). However, my rotation is 90 degrees, facing east. I need to adjust the rotation to center the target within its field of view. \newline
2) Obstacles don't affect rotation. \newline
3) RotateRight \\
\bottomrule
\end{tabular}
\caption{Response templates we used to build CoT explanation traces for three common rotation scenarios.}
\label{tab:response_template_rotate}
\end{table*}

\subsection{Data Postprocessing}
\label{app:data_postprocessing}
Directly using all actions in every trajectory to conduct imitation learning brings about an extremely imbalanced dataset. As shown in \cref{tab:data_postprocessing}, 77.94\% actions are \moveahead. Then we downsampled \moveahead actions in the instruction dataset. We only retain 25\% of the \moveahead actions resulting in a more balanced dataset.

Then, we also remove conflicting data. We find that steps from different trajectories within the same house occasionally exhibit conflicting information. They have the exact same input information but different action predictions. This happens when two overlapped trajectories diverge at some point due to different target objects. Those conflicting data can confuse the fine-tuned LVLM and lead to worse performance. Thus, we remove those conflicting data from our dataset.

We show the final distribution of our dataset in \cref{tab:data_postprocessing} (w/ Postproc).

\section{Supplementary Experiment Details}
In this section, we provide supplementary experiment results and show the prompt details.
\subsection{Ablation Study Prompt}
\label{app:ablation_stduy}
We change the prompt templates for tuning our agent \modelname in the ablation studies. 

(1) For adding the gold label of position difference ($\diamond$ w/o ET \& w Gold and $\diamond$ w Gold Label), we append a new sentence ``The difference to the target object is \textbf{[position\_diff]}'' to the end of the \textbf{Episode-Specific Information} part of the instruction template. 

(2) For removing the explanation traces, the \textbf{Prediction Steps} part of the instruction template is replaced with one shorter sentence, ``Please generate the next step given the above states.'' 

(3) For the \textbf{difference-equation prompt} in the $\diamond$ w Diff-EQ experiment, we replace \textbf{[position\_diff]} in all explanation traces with the equation: \textbf{\textbf{[grid\_obj\_pos]}} - \textbf{[agent\_pos]} = \textbf{[position\_diff]}.

\subsection{Full Results of the Image Number Experiment}
\label{app:image_number}
We provide the full results of the image number experiment of all metrics in \cref{tab:image_number}. Besides the default prompt structure, we also conduct experiments with the difference-equation prompt, which is introduced as ``$\diamond$ w Diff-EQ'' in the ablation study. 

\begin{table*}[t!]
\centering
\small

\begin{subtable}[t]{\textwidth}
\setlength{\tabcolsep}{4.9pt}
\centering
\begin{tabular}{l||ccc|ccc|ccc}
\toprule
\multirow{2}{*}{\textbf{Number}}&\multicolumn{3}{c|}{\textbf{Valid}} &\multicolumn{3}{c|}{\textbf{Test}}&\multicolumn{3}{c}{\textbf{All}}\\
&\textbf{SR}&\textbf{SPL} &\textbf{SEL} &\textbf{SR}&\textbf{SPL} &\textbf{SEL} & \textbf{SR}&\textbf{SPL}&\textbf{SEL} \\
    \midrule
2 images&50.00&41.64&40.23&50.00&40.71&41.03&50.00&41.17&40.63\\
4 images&\textbf{57.41}&\textbf{47.84}&\textbf{47.90}&\textbf{54.94}&\textbf{44.45}&45.83&\textbf{56.17}&\textbf{46.15}&\textbf{46.86}\\
6 images&45.37&37.61&35.37&50.31&40.88&38.70&47.84&39.25&37.03\\
8 images&49.07&40.82&41.52&54.01&43.61&\textbf{46.22}&51.54&42.22&43.87\\
\bottomrule
\end{tabular}
\caption{Performance of the default prompt.}
\end{subtable}

\begin{subtable}[t]{\textwidth}
    \setlength{\tabcolsep}{4.9pt}
\centering
\begin{tabular}{c||ccc|ccc|ccc}
	\toprule
        \multirow{2}{*}{\textbf{Number}}&\multicolumn{3}{c|}{\textbf{Valid}} &\multicolumn{3}{c|}{\textbf{Test}}&\multicolumn{3}{c}{\textbf{All}}\\
	&\textbf{SR}&\textbf{SPL} &\textbf{SEL} &\textbf{SR}&\textbf{SPL} &\textbf{SEL} & \textbf{SR}&\textbf{SPL}&\textbf{SEL} \\
            \midrule
 2 images&46.30&38.54&38.93&52.16&42.13&44.76&49.23&40.34&41.84\\
 4 images&54.63&45.02&\textbf{46.48}&54.94&44.04&\textbf{47.40}&54.78&44.53&\textbf{46.94}\\
 6 images&49.07&40.82&38.88&51.23&41.56&40.20&50.15&41.19&39.54\\ 
 8 images&\textbf{55.56}&\textbf{45.81}&43.40&\textbf{57.41}&\textbf{46.39}&44.40&\textbf{56.48}&\textbf{46.10}&43.90\\
		\bottomrule
	\end{tabular}
\caption{Performance of the difference-equation prompt.}
\end{subtable}
\caption{The investigation of the hyperparameter: image number. We provide different numbers of recent visual observations of the embodied agent. Besides the default prompt we use, we also evaluate the difference-equation prompt. We bold the best performance.}
\label{tab:image_number}
\end{table*}

\subsection{Full Results of the Action Number Experiment}
\label{app:action_number}
We provide the full results of the action number experiment with the default prompt in \cref{tab:full_action_number}. We also provide the results with the difference-equation prompt in \cref{tab:action_number_eq_diff_prompt}. The prompt is used as ``$\diamond$ w Diff-EQ'' in the ablation study. 

\begin{table*}[t]
\centering
\small
\setlength{\tabcolsep}{4.9pt}
\begin{tabular}{c||ccc|ccc|ccc}
	\toprule
        \multirow{2}{*}{\textbf{Number}}&\multicolumn{3}{c|}{\textbf{Valid}} &\multicolumn{3}{c|}{\textbf{Test}}&\multicolumn{3}{c}{\textbf{All}}\\
	&\textbf{SR}&\textbf{SPL} &\textbf{SEL} &\textbf{SR}&\textbf{SPL} &\textbf{SEL} & \textbf{SR}&\textbf{SPL}&\textbf{SEL} \\
            \midrule
4 steps&46.30&38.66&39.82&45.99&37.45&39.60&46.14&38.05&39.71\\
8 steps&\textbf{57.41}&\textbf{47.84}&\textbf{47.90}&\textbf{54.94}&\textbf{44.45}&\textbf{45.83}&\textbf{56.17}&\textbf{46.15}&\textbf{46.86} \\
12 steps&42.59&35.92&36.43&50.93&41.15&43.20&46.76&38.53&39.81\\
16 steps&51.85&43.01&42.72&53.40&43.03&44.08&52.62&43.02&43.40\\
		\bottomrule
\end{tabular}
\caption{Hyperparameter investigation. Our agents receive various numbers of recent actions and positions within the default prompt.}
\label{tab:full_action_number}
\end{table*}

\begin{table*}[!t]
\centering
\small
\setlength{\tabcolsep}{4.9pt}
\begin{tabular}{c||ccc|ccc|ccc}
	\toprule
        \multirow{2}{*}{\textbf{Number}}&\multicolumn{3}{c|}{\textbf{Valid}} &\multicolumn{3}{c|}{\textbf{Test}}&\multicolumn{3}{c}{\textbf{All}}\\
	&\textbf{SR}&\textbf{SPL} &\textbf{SEL} &\textbf{SR}&\textbf{SPL} &\textbf{SEL} & \textbf{SR}&\textbf{SPL}&\textbf{SEL} \\
            \midrule
4 steps&45.37&37.76&40.04&52.47&42.87&46.45&48.92&40.31&43.25\\
8 steps&54.63&45.02&46.48&54.94&44.04&47.40&54.78&44.53&46.94\\
12 steps&51.85&42.80&44.06&\textbf{60.19}&\textbf{48.10}&\textbf{51.63}&56.02&45.45&47.84\\
16 steps&\textbf{56.48}&\textbf{46.39}&\textbf{47.89}&59.26&47.52&50.82&\textbf{57.87}&\textbf{46.95}&\textbf{49.36}\\
		\bottomrule
	\end{tabular}
\caption{Hyperparameter investigation. We provide different numbers of recent actions and positions to the embodied agent. We use the \textbf{difference-equation prompt} in this table. The best performance is bolded.}
\label{tab:action_number_eq_diff_prompt}
\end{table*}

\subsection{Full Results of the Few-Shot Experiment}
\label{app:few_shot}
We provide the full results of the few-shot learning with the default prompt in \cref{tab:full_few_shot_experiment}. We also provide the results with the difference-equation prompt in \cref{tab:few_shot_eq_diff_prompt}. The prompt is used as ``$\diamond$ w Diff-EQ'' in the ablation study.

\begin{table*}[t!]
\centering
\small
\setlength{\tabcolsep}{4.9pt}
\begin{tabular}{c||ccc|ccc|ccc}
	\toprule
        \multirow{2}{*}{\textbf{\% Data}}&\multicolumn{3}{c|}{\textbf{Valid}} &\multicolumn{3}{c|}{\textbf{Test}}&\multicolumn{3}{c}{\textbf{Test Diff w GPT-4o}}\\
	&\textbf{SR}&\textbf{SPL} &\textbf{SEL} &\textbf{SR}&\textbf{SPL} &\textbf{SEL} & \textbf{$\Delta_{SR}$} & \textbf{$\Delta_{SPL}$} & \textbf{$\Delta_{SEL}$}  \\
            \midrule
20&37.04&30.86&30.22&38.89&31.40&32.13&$\uparrow$0.62&$\downarrow$0.34&$\uparrow$4.21\\
40&33.33&27.95&27.13&38.27&31.00&30.79&$\downarrow$0.00&$\downarrow$0.74&$\uparrow$2.87\\
60&49.07&40.68&42.01&52.12&42.25&44.40&$\uparrow$13.85&$\uparrow$10.51&$\uparrow$16.48\\
80&50.00&41.46&43.36&49.69&40.26&42.49&$\uparrow$11.42&$\uparrow$8.52&$\uparrow$14.57\\
100&\textbf{57.41}&\textbf{47.84}&\textbf{47.90}&\textbf{54.94}&\textbf{44.45}&\textbf{45.83}&$\uparrow$16.67&$\uparrow$12.71&$\uparrow$17.91\\
\bottomrule
\end{tabular}
\caption{Few-shot learning ability. We test our agent with different proportions of training data with the default prompt.}
\label{tab:full_few_shot_experiment}
\end{table*}

\begin{table*}[t!]
\centering
\small
\setlength{\tabcolsep}{4.9pt}

\begin{tabular}{c||ccc|ccc|ccc}
	\toprule
        \multirow{2}{*}{\textbf{\% Data}}&\multicolumn{3}{c|}{\textbf{Valid}} &\multicolumn{3}{c|}{\textbf{Test}}&\multicolumn{3}{c}{\textbf{Test Diff w GPT-4o}}\\
	&\textbf{SR}&\textbf{SPL} &\textbf{SEL} &\textbf{SR}&\textbf{SPL} &\textbf{SEL} & \textbf{$\Delta_{SR}$} & \textbf{$\Delta_{SPL}$} & \textbf{$\Delta_{SEL}$}  \\
            \midrule
20&37.96&32.01&36.72&32.10&26.45&31.15&$\downarrow$6.17&$\downarrow$5.29&$\uparrow$3.23\\
40&38.89&32.76&38.14&33.33&27.07&31.94&$\downarrow$4.94&$\downarrow$4.67&$\uparrow$4.02\\
60&\textbf{62.96}&\textbf{51.47}&\textbf{52.19}&\textbf{58.95}&\textbf{47.34}&\textbf{49.53}&$\uparrow$20.68&$\uparrow$15.60&$\uparrow$21.61\\
80&57.41&46.79&45.73&57.10&45.98&46.58&$\uparrow$18.83&$\uparrow$14.24&$\uparrow$18.66\\
100&54.63&45.02&46.48&54.94&44.04&47.40&$\uparrow$16.67&$\uparrow$12.30&$\uparrow$19.48\\
		\bottomrule
\end{tabular}
\caption{The evaluation of the few-shot learning ability of our agent \modelname. We test our agent with different proportions of sampled episodes in each room. We compare the results with the GPT-4o and test the \textbf{difference-equation prompt} in this table.}
\label{tab:few_shot_eq_diff_prompt}
\end{table*}

\subsection{Zero-shot Transferring Evaluation}
\label{app:zero_shot_transfer}
iTHOR and ProcTHOR both encompass four distinct scene types: bedrooms, living rooms, kitchens, and bathrooms. iTHOR has 30 rooms for each scene type, designed by professional 3D artists. Meanwhile, ProcTHOR is a procedural house-generation system that constructs 10,000 unique houses automatically. Similar to iTHOR, we sample 30 houses for each scene type from ProcTHOR. Four episodes are sampled in each house to evaluate our agent.

On the HM3D dataset, we test \modelname on all houses. We compare our model with a lot of baselines on the HM3D dataset, including two lines of work: fine-tuned methods and zero-shot methods. As shown in \cref{tab:hm3d_score}, the SPL scores of baselines are mostly lower than 30\%. In contrast, the SPL score of \modelname is 34.11\%, 10 points higher than theirs. 

Meanwhile, we want to emphasize that those LLM or VLM-based complicated navigation framework~\cite{ramrakhya2023pirlnav, majumdar2022zson, zhou2023esc, yokoyama2024vlfm, yin2024sg, long2024instructnav} usually involve a lot more information than \modelname. For example, our method only takes ego-centric RGB images as input, while those methods also get depth information, scene graphs, local policy navigation tools, extra VLMs (like GPT-4) besides backbone VLM, semantic segmentation, and panoramic field of views. This makes the comparison \textbf{inherently unfair}, as their methods rely on significantly more resources and additional modalities.

Meanwhile, we also \textbf{do not include} those complicated frameworks in other experiments due to the following reasoning: (1) They fall outside the evaluation scope of the current LVLMs; (2) They utilize different simulation platforms, such as Habitat~\cite{szot2021habitat}, rather than AI2THOR, which introduces significant compatibility issue with AI2THOR; (3) As abovementioned, they also introduce a lot of extra information beyond than ego-centric RGB, making the evaluation unfair.

\begin{table*}[t]
\centering
\small
\begin{tabular}{l|c|c}
\toprule
\textbf{Methods} & \textbf{SPL} & \textbf{SR} \\
\midrule
\multicolumn{3}{c}{\textbf{Fine-tuned}} \\
ProcTHOR~\cite{deitke2022️} & 31.8 & 54.4 \\
PIRLNav~\cite{ramrakhya2023pirlnav} & 27.1 & 64.1 \\
\midrule
\multicolumn{3}{c}{\textbf{Zero-Shot}} \\
ProcTHOR (zero-shot) & 7.7 & 13.2 \\
ZSON~\cite{majumdar2022zson} & 12.6 & 25.5 \\
ESC~\cite{zhou2023esc} & 22.3 & 39.2 \\
VLFM~\cite{yokoyama2024vlfm} & 30.4 & 52.5 \\
SG-Nav~\cite{yin2024sg} & 24.8 & 53.9 \\
InstructNav~\cite{long2024instructnav} & 20.9 & \textbf{58.0} \\
\modelname (Our) & \textbf{34.1} & 41.5 \\
\bottomrule
\end{tabular}
\caption{Comparison of different models on HM3D, including fine-tuned and zero-shot methods.}
\label{tab:hm3d_score}
\end{table*}

\section{Implementation Details}
\label{app:implemntation}
We train our agent \modelname from Idefics 2 on 8 NVIDIA A100 GPUs using the Megation-LM framework~\citep{megatron-lm}. All parts of the Idefices 2 are fine-tuned, including the LLM, vision encoder, and modality projector. We load the model in BF16 and fine-tune it for one epoch with the learning rate and batch size of 2e-5 and 64, respectively. The best checkpoint is selected according to the sum of all metrics on the validation set. The image size sampled from the AI2THOR is 300 $\times$ 300. For the baselines, we use the same instructions as our agent and ask them to predict the next action directly. Similarly, we also provide the history of the recent 8 steps (i.e., actions and agent positions) and the visual observation of the recent 4 steps. The exceptions are blind LLMs and Llava 1.5, which can handle zero and one image, respectively. We access the closed-source LVLMs via the OpenAI API\footnote{https://platform.openai.com/docs/api-reference} with the specific versions of \texttt{gpt-4-vision-preview} and \texttt{gpt-4o-2024-08-06}.

Here, we discuss the computational cost. For training, we use 8 $\times$ A100 GPUs to train our model. After adding the CoT traces, the training time increases from 10 hours to 17 hours, which is acceptable. Meanwhile, we use 1 A100 GPU to deploy the model. The generation time increases from 0.28s to 1.03s per query when we add the CoT traces. In practice, we can use more GPUs to speed up the inference. For example, RT-2~\cite{brohan2023rt} uses a multi-TPU cloud service to achieve a frequency of 5Hz for a 5B model and 3Hz for a 55B model. The computational cost, both in training and inference, is reasonable in our experiments.

\section{More House Examples}
\label{app:more_house_example}
\begin{figure*}[t]
    \centering
    \begin{subfigure}[b]{0.3\textwidth}
        \centering
        \includegraphics[width=\textwidth]{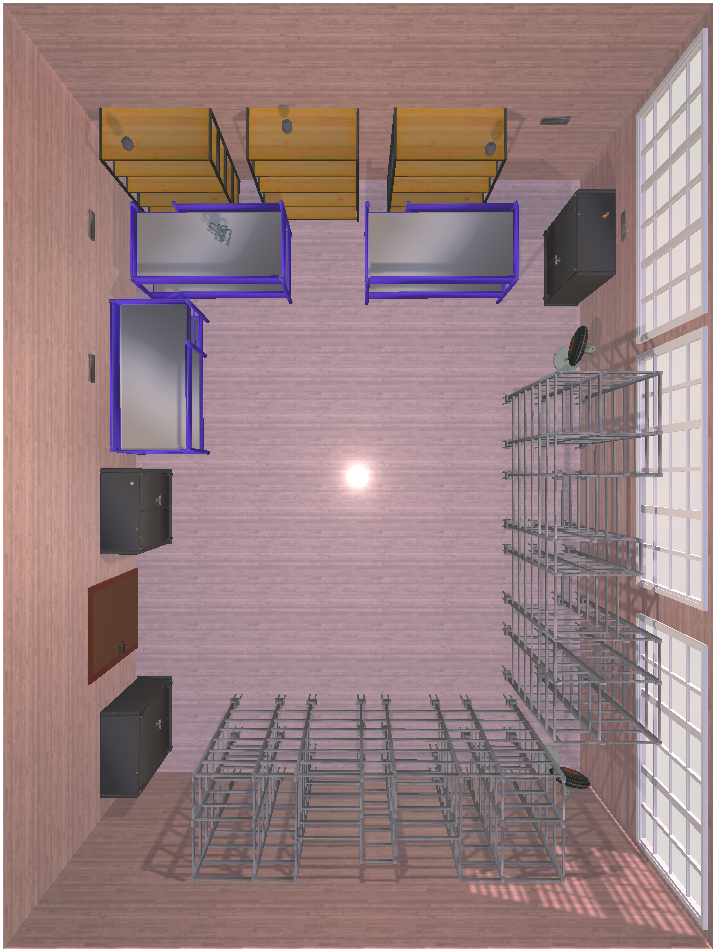}
        \caption{a warehouse with large windows}
    \end{subfigure}
    \hfill
    \begin{subfigure}[b]{0.3\textwidth}
        \centering
        \includegraphics[width=\textwidth]{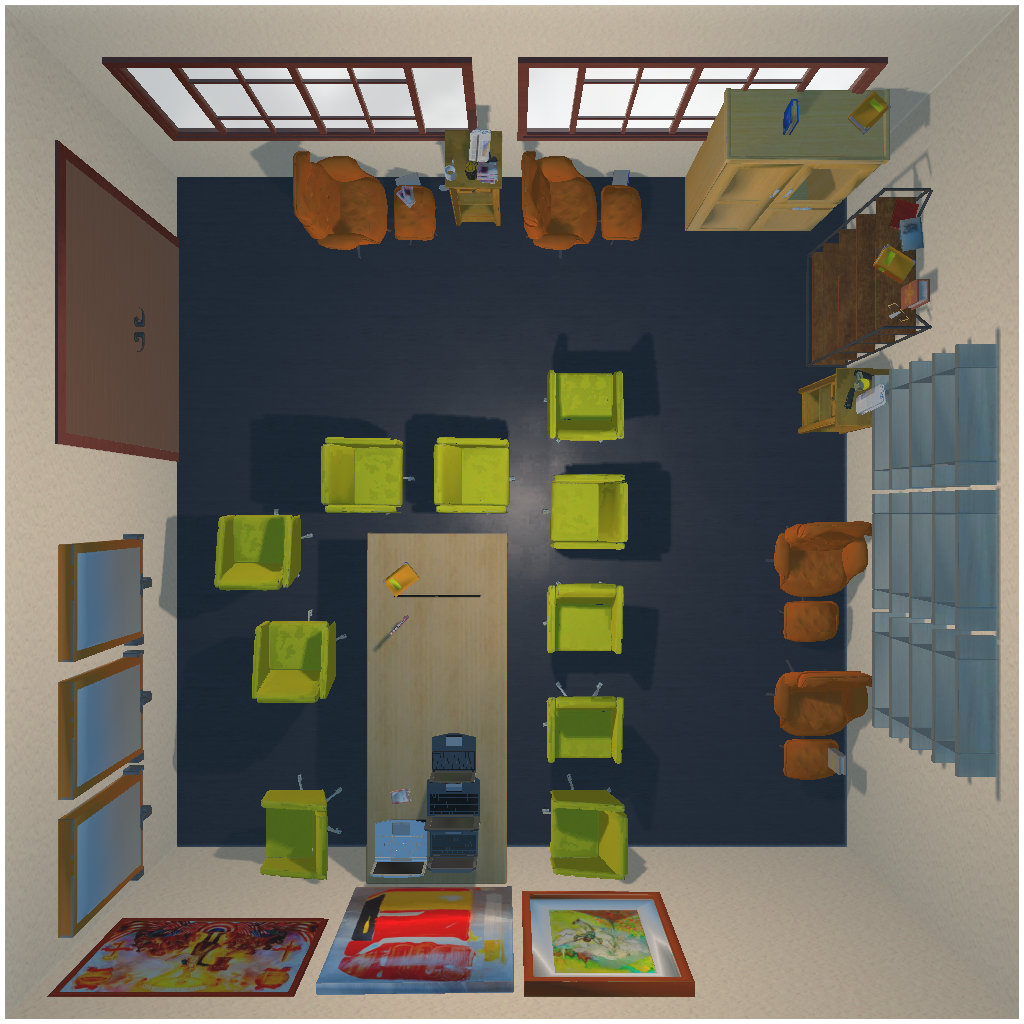}
        \caption{a meeting room with artistic paintings}
    \end{subfigure}
    \hfill
    \begin{subfigure}[b]{0.3\textwidth}
        \centering
        \includegraphics[width=\textwidth]{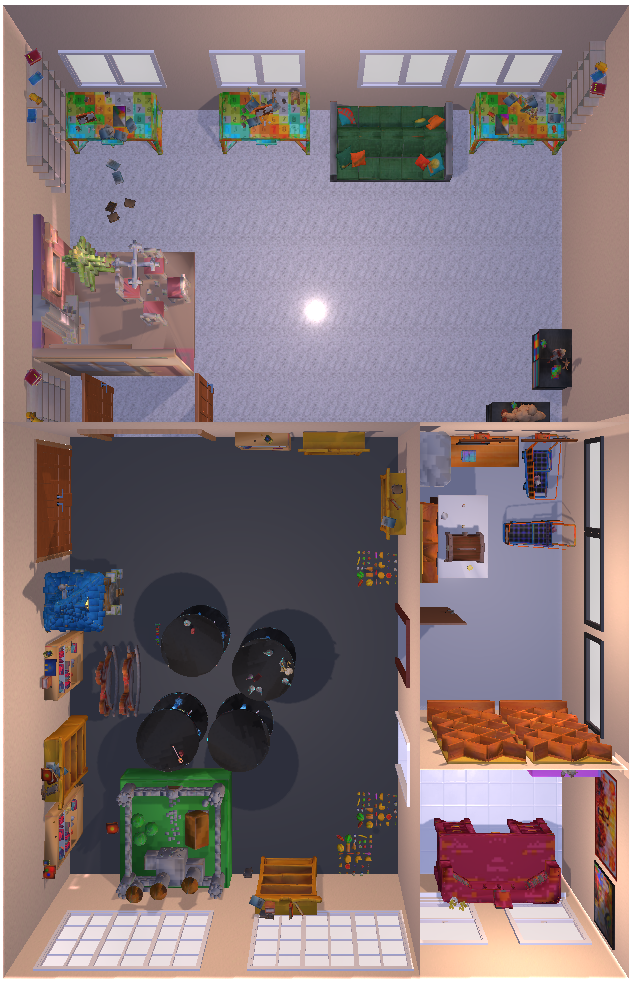}
        \caption{a toy store with a magical kingdom theme.}
    \end{subfigure}
    
    \vspace{1em}
    
    \begin{subfigure}[b]{0.3\textwidth}
        \centering
        \includegraphics[width=\textwidth]{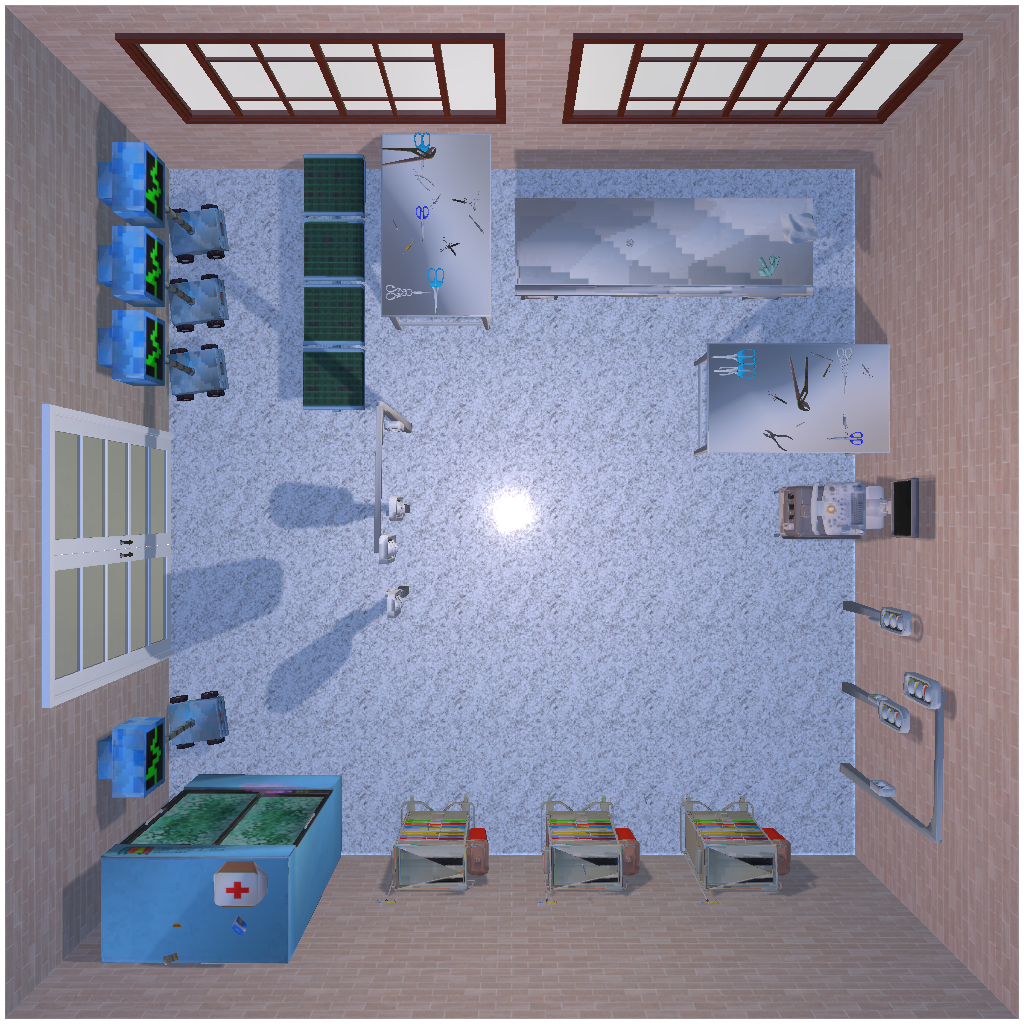}
        \caption{an operating room used by surgeons and nurses featuring light-colored tiles on the walls.}
    \end{subfigure}
    \hfill
    \begin{subfigure}[b]{0.3\textwidth}
        \centering
        \includegraphics[width=\textwidth]{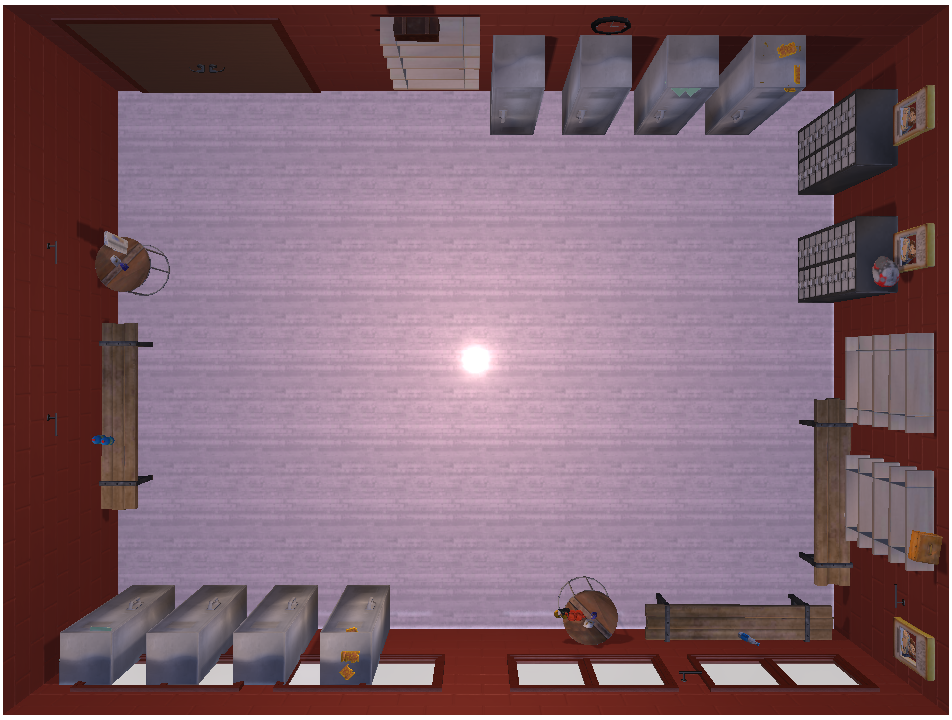}
        \caption{an industrial locker room}
    \end{subfigure}
    \hfill
    \begin{subfigure}[b]{0.3\textwidth}
        \centering
        \includegraphics[width=\textwidth]{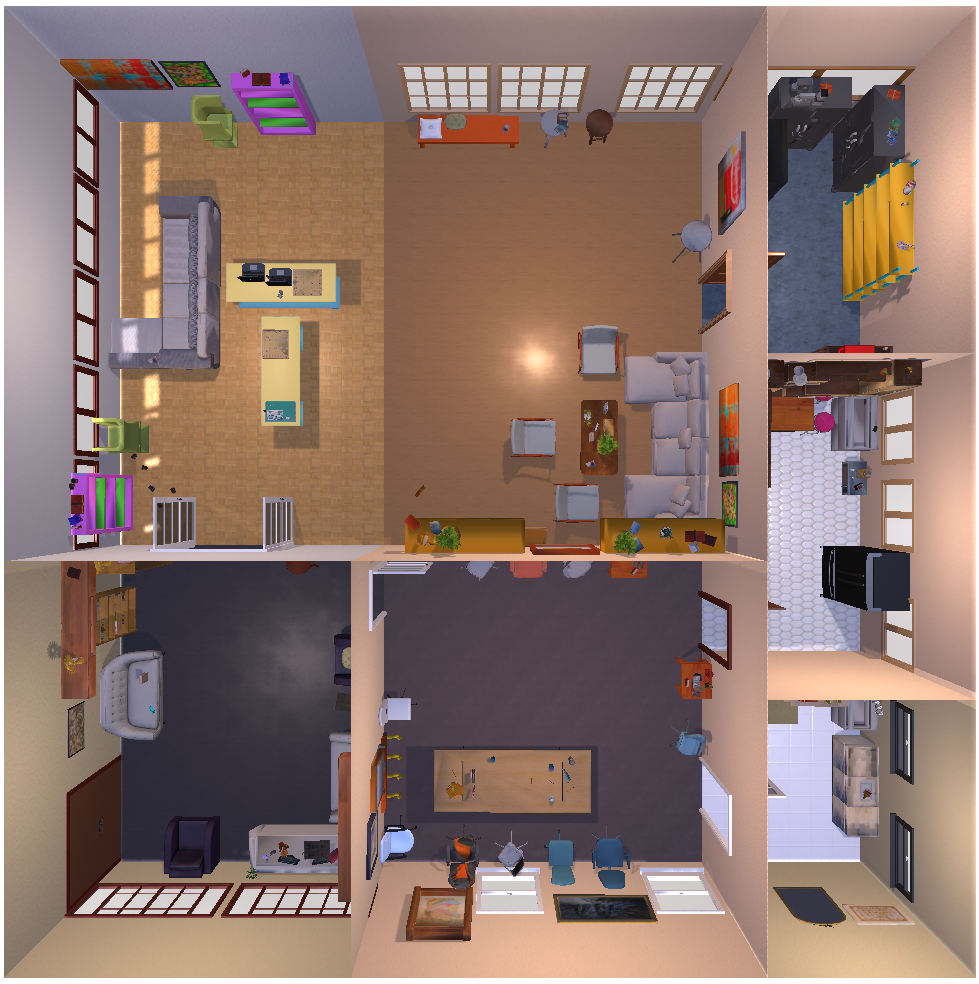}
        \caption{a community center with a versatile and inclusive theme featuring a spacious room size.}
    \end{subfigure}
    
    \caption{Examples of different houses in \scenedataname.}
    \label{fig:more_house_example}
\end{figure*}

In this section, we present more houses built in our \scenedataname dataset in \cref{fig:more_house_example}.

\begin{figure*}[t]
    \centering
    \begin{subfigure}[b]{0.37\textwidth}
        \centering
        \includegraphics[width=\textwidth]{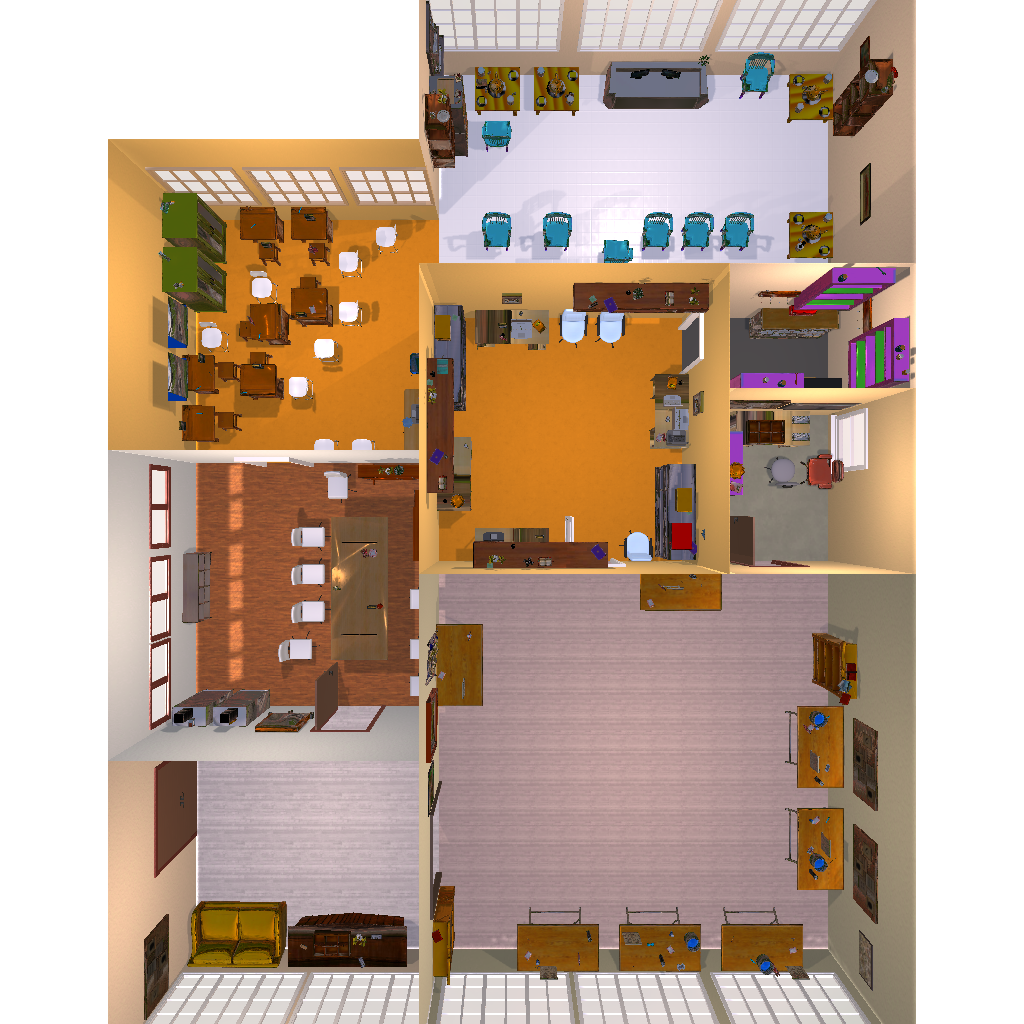}
        \caption{a community center with versatile flooring}
    \end{subfigure}
    \hfill
    \begin{subfigure}[b]{0.45\textwidth}
        \centering
        \includegraphics[width=\textwidth]{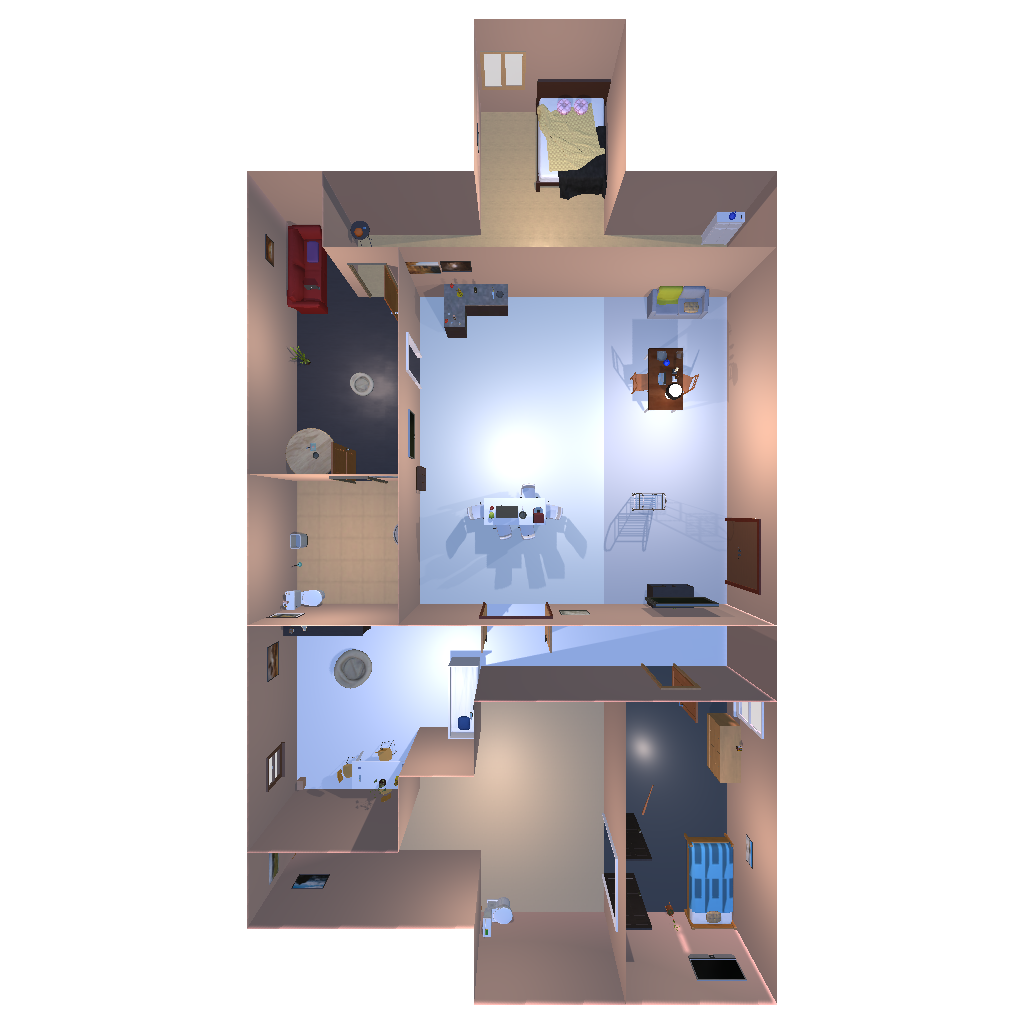}
        \caption{a house from ProcTHOR}
    \end{subfigure}
    \caption{Comparing our houses with ProcTHOR.}
    \label{fig:house_comparison}
\end{figure*}

\subsection{Comparison with ProcTHOR}
In \cref{fig:house_comparison}, we compare houses from our dataset and ProcTHOR with the same number of rooms (8 rooms). Obviously, our scene is more complex with more objects. Quantitively, our scene contains 466 objects, and the scene from ProcTHOR contains only 74 objects.

\end{document}